\documentclass[runningheads]{llncs}
\usepackage[T1]{fontenc}
\usepackage{graphicx}
\usepackage[misc]{ifsym}

\usepackage{accvabbrv}

\usepackage{graphicx}
\usepackage{booktabs}

\usepackage[accsupp]{axessibility}  
\usepackage{hyperref}

\usepackage{arydshln}
\usepackage{amssymb}
\usepackage{pifont}
\usepackage{microtype}
\usepackage[capitalize]{cleveref}

\newcommand{\comment} [1]{} 

\def\ssp{\hspace{3pt}}
\def\msp{\hspace{6pt}}

\def\l1{\ensuremath{\ell_1}\xspace}
\def\l2{\ensuremath{\ell_2}\xspace}

\makeatletter
\DeclareRobustCommand\onedot{\futurelet\@let@token\@onedot}
\def\@onedot{\ifx\@let@token.\else.\null\fi\xspace}

\def\ie{\emph{i.e}\onedot}

\makeatother

\newcommand{\cmark}{\ding{51}}%
\newcommand{\xmark}{\ding{55}}%

\newcommand{\method}{CBNC\xspace}

\usepackage{color}

\begin{document}
\title{Co-Segmentation without any Pixel-level Supervision with Application to Large-Scale Sketch Classification}
\titlerunning{Co-Segmentation without any Pixel-level Supervision}
%
\author{Nikolaos-Antonios Ypsilantis $^{\text{\Letter}}$ \orcidID{0000-0002-3322-6925} \\ \and 
 Ond\v{r}ej Chum\orcidID{0000-0001-7042-1810}}
\authorrunning{N.-A. Ypsilantis and  O. Chum}
%
\institute{Visual Recognition Group, FEE, Czech Technical University in Prague
\\
\Letter \email{ ypsilnik@fel.cvut.cz}
}
\maketitle              
%


\begin{abstract}
    This work proposes a novel method for object co-segmentation, i.e. pixel-level localization of a common object in a set of images, that uses no pixel-level supervision for training. 
    Two pre-trained Vision Transformer (ViT) models are exploited: ImageNet classification-trained ViT, whose features are used to estimate rough object localization through intra-class token relevance, and a self-supervised DINO-ViT for intra-image token relevance.
    On recent challenging benchmarks, the method achieves state-of-the-art performance among methods trained with the same level of supervision (image labels) while being competitive with methods trained with pixel-level supervision (binary masks).
    
    The benefits of the proposed co-segmentation method are further demonstrated in the task of large-scale sketch recognition, that is, the classification of sketches into a wide range of categories.
    The limited amount of hand-drawn sketch training data is leveraged by exploiting readily available image-level-annotated datasets of natural images containing a large number of classes.
    To bridge the domain gap, the classifier is trained on a sketch-like proxy domain derived from edges detected on natural images.
    We show that sketch recognition significantly benefits when the classifier is trained on sketch-like structures extracted from the co-segmented area rather than from the full image.
    
    Code: \texttt{https://github.com/nikosips/CBNC}.
\end{abstract}
%
\section{Introduction}
\begin{figure}[t]
\centering
\scalebox{0.87}{
\begin{tabular}{@{\ssp}c@{\ssp}c@{\ssp}c@{\ssp}c@{\ssp}c@{\ssp}c}
\includegraphics[height=2.15cm, width=2.15cm]{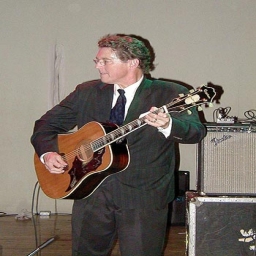}
&
\includegraphics[height=2.15cm, width=2.15cm]{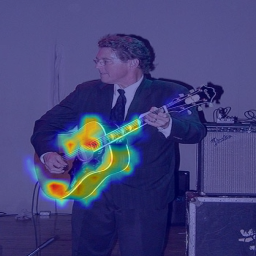}
&
\includegraphics[height=2.15cm, width=2.15cm]{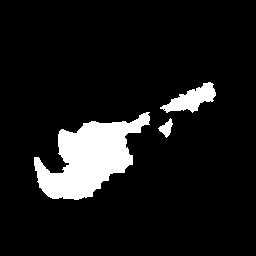}
&
\includegraphics[height=2.15cm, width=2.15cm]{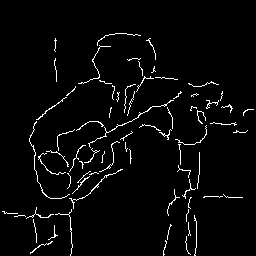}
&
\includegraphics[height=2.15cm, width=2.15cm]{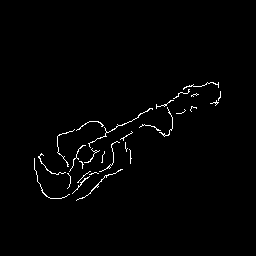}
&
\includegraphics[height=2.15cm, width=2.15cm]{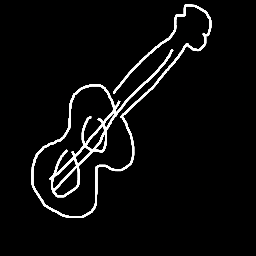} \\
(a) & (b) & (c) & (d) & (e) & (f) \\
\end{tabular}
}
\caption{ (a) Original RGB image of the class ``guitar''; (b) Patch-level class relevance based on inter-image ImageNet ViT token similarity (low values in blue, high values in red); (c) Patch-level segmentation based on DINO-ViT intra-image token similarity with bias from the class relevance, followed by refinement via GrabCut; 
(d) training example for sketch recognition training extracted from the whole image as in~\cite{etc22},
(e) from the object mask; 
(f) test time sketch fed to the sketch classifier.
We show that training with examples like (e) instead of (d) improves the performance of sketch classifiers trained without sketches.
}
\label{fig:motivation}
\end{figure}

We consider the problem of discovering and segmenting the common object out of a set of images, given no information other than the images themselves.
Known as object co-segmentation or co-salient object detection, the goal is to obtain the pixel-level extent of all objects of a class that is repeated in a set of images.
The difficulty of the task depends on the intra-class variation of the instances shown in the set of images, the complexity of a background, and the presence of other salient object classes.
It constitutes the first step in numerous applications, including collection-aware crops~\cite{jgs10}, image retrieval~\cite{cmh+14}, image quality assessment~\cite{wly+19}, and weakly supervised learning~\cite{zzl+19}.
In this paper, we introduce a novel application of co-segmentation to background removal in large-scale sketch classification trained from photographs; see overview in Figure~\ref{fig:motivation}.

Existing work in the co-segmentation field focuses mostly on developing different training losses and architectures~\cite{zhl+21,sds+22,fff+21,yxz+22}.
While these methods achieve impressive results, they rely on supervised ground truth segmentation masks, which take huge amounts of time and effort to obtain through human annotation, particularly for large-scale datasets. 
We focus on developing a co-segmentation pipeline that does not require pixel-level annotation as supervision in training. 

Vision Transformers (ViTs) have recently shown great performance in a number of computer vision tasks. 
The capacity of the ViT models is large, and different training approaches with the same backbone architecture result in significantly different properties. 
The ViT trained with the self-supervised approach DINO~\cite{ctm+21} has been shown to exhibit salient foreground object aware local (patch) features, which have been exploited in many works that focus on segmentation tasks, like salient object discovery and foreground segmentation~\cite{wsh+22,agb+21,mrl+22}.
However, we observe that these features can not be directly used for the task of co-segmentation, as they are incapable of accurate inter-image category awareness.

We propose a method for the co-segmentation task, called Class-relevance Biased N-Cut (CBNC), which exploits two pre-trained ViTs, one pre-trained for classification on ImageNet and one pre-trained with the self-supervised DINO objective, effectively taking advantage of the best of both worlds.
In the first step, the patch-level class relevance is estimated from the set of class images (Figure~\ref{fig:motivation} (b)).
For this step, the inter-image similarity of features obtained from ViT~\cite{dbk+20} trained with ImageNet~\cite{dsl+09} supervision is used.
In the second step, the per-image patch-level segmentation is obtained,
by using Biased N-Cuts~\cite{mvm11} that are guided by the class relevance bias and use the intra-image similarity of DINO-ViT~\cite{ctm+21} features (Figure~\ref{fig:motivation} (c)).

Furthermore, we introduce a novel application of salient object co-segmentation by demonstrating the power of the proposed method on the problem of large-scale sketch recognition trained on image-level annotated photographs.
The difficulty of sketch recognition grows with the increasing number of categories that need to be recognized. Classical supervised approaches do not scale simply because there is not enough training data.
Extrapolating the information available from the Sketchy~\cite{sbh+16} dataset construction, training examples for over 1000 categories would require tens of man-years of sketch drawing. To address the scalability issue, the task of cross-domain photograph-to-sketch classification was recently proposed in~\cite{etc22}.

The domain gap between photographs (training data) and sketches (test data) is overcome by transforming images in an intermediate domain, called \emph{randomized Binary Thin Edges} (rBTE), in \cite{etc22}.
This sketch-like domain is constructed from edgemaps of natural images -- \emph{using the entire image} -- that undergo a set of randomized augmentations to resemble sketches. 
For clean images containing only the object of interest and no or little background, rBTEs often capture the shape (provided by the outline edges) and visually important features (given by the inside edges). However, in the presence of cluttered background or co-occurring objects, the rBTEs contain a large number of irrelevant edges (for examples, see Figure ~\ref{fig:motivation}). In such a case, \cite{etc22} relies on the ability of the network to select the common structure in rBTEs extracted from multiple training photographs. 

We propose to first segment the object of interest using the proposed method and only then use the rBTE pseudo-domain to train the sketch classifier. 
The difference in the training data passed to the classifier is illustrated in Figure~\ref{fig:motivation} (d) for the full images as in \cite{etc22} and (e) for our proposed approach.

The contribution of this work is twofold.
\begin{itemize}
    \item First, we propose a novel co-salient object discovery and co-segmentation method that outperforms state-of-the-art methods on standard benchmarks that use the same level of supervision, where the performance is measured in terms of similarity to the ground truth segmentations. 
    \item Second, we propose to train a sketch classifier from rBTEs constructed only from automatically discovered regions corresponding to the object of interest. 
    We experimentally show (i) that eliminating the background clutter significantly improves the sketch recognition performance and (ii) our proposed method outperforms other image-level-supervised methods in this task, measured in the recognition performance.
\end{itemize}

\section{Related work}
In this section, we briefly review relevant work to the tasks of object discovery, co-segmentation and sketch classification and retrieval, as well as to Vision Transformers.
\paragraph{Object discovery and segmentation.}
Object discovery refers to the task of localizing the main unknown object of an image.
The seminal work on N-cuts~\cite{sma97} provided a way of solving the problem of object discovery given only an image using spectral graph theory.
It represents the image by a graph whose nodes are the pixels of the image, weighted by pixel similarities based on color features and performs graph clustering to produce a binary mask of the main object.
N-cuts on DINO ViT features are used in~\cite{wsh+22},
giving higher-level semantic information to guide the segmentation.
DINO ViT features are also used in object discovery without N-Cuts in ~\cite{spv+21}.
Spectral graph theory is also examined for the task of unsupervised object detection and localization using self-supervised features in~\cite{sax22,mrl+22}.
In the case of many salient objects in an image, these methods require guidance to focus on objects of interest.
Biased N-cut~\cite{mvm11}, which was proposed in order to guide the segmentation by user-defined seeds in the image, provides a way to guide the segmentation given initial coarse guesses.
\paragraph{Object co-segmentation.}
The task of object co-segmentation~\cite{rother2006cosegmentation} and the similar task of co-salient detection aim to localize the common salient object in a set of images.
Different methods have been proposed to solve this task, which can be categorized depending on the level of supervision they need.
Many recent approaches based on deep learning use large amounts of pixel-level annotations~\cite{zhl+21,sds+22,fff+21,yxz+22}. 
In~\cite{wzw+19}, the method Deep Descriptor Transforming (DDT) was proposed for local features obtained by Convolutional Neural Networks.
It proposed that the problem of object co-localization can be tackled by leveraging correlations between the set of all local features of images containing a common object. 
Those local features were extracted by pre-trained CNNs, showing the importance of good model reuse.
A positive correlation of local features with the principal direction would act as a detector of the common object.
Our proposed co-segmentation method builds on this work, with the CNN features being replaced by Vision Transformer features.
The method proposed in~\cite{agb+21} also uses Vision Transformer features to create a co-segmentation pipeline, in a different way than we do.
More specifically, they propose to first cluster the set of all local features of the class into semantic clusters that induce segments in all images.
A voting procedure is then used to select those that are common in all images and are salient in order to create the co-segmentation masks.
In the recent work~\cite{sxa22}, the task of language-guided co-segmentation is introduced, where a textual description of the class is used as an additional input in order to perform the co-segmentation.

\paragraph{Sketch recognition.}
In~\cite{etc22}, scaling up the number of sketch classes is tackled by training the sketch classification network using image-level annotated photographs, while the classifier is evaluated on sketch data. 
A proxy domain called \emph{randomized Binary Thin Edges} is introduced to bridge the train-test domain gap. 
Natural images are mapped to the rBTE domain by the following steps: first, the image is transformed to an edge map, represented as an edge probability map, by a randomly selected edge detector, specifically one of Structured Forests \cite{dz13}, HED \cite{sz17} and BCDN \cite{jsm+19}.
Subsequently, the edge map undergoes a random geometric augmentation, and then is thinned using Non-Maxima suppression to resemble the thin nature of sketches.
Hysteresis thresholding is then applied to binarize the edgemap, and finally, very small connected components are discarded. 
In our sketch recognition application, we build upon~\cite{etc22}. Specifically, we use an identical training pipeline, but instead of using the rBTE from the entire image, only the area segmented by the proposed algorithm is used. 
Single domain generalization methods, such as~\cite{SelfReg,Wlq+21,SagNet}, are also directly applicable to the task of training on photographs and evaluating on sketches. However, it has already been experimentally evaluated in~\cite{etc22} that generic domain generalization approaches do not perform as well as methods designed to train on natural image domain and test on sketches.
\paragraph{Sketch Based Image Retrieval.}
Edge detection on natural images was exploited by pre-neural-network approaches to sketch-based image retrieval and matching with hand-crafted descriptors~\cite{rc13,bc15,sbo15,pm14,tc17}.
Modern methods based on deep learning commonly use a two-branch architecture~\cite{brp+16b,sbh+16,yls+16,sdm17}.
One branch is used for natural images, the other for sketches, both branches learn to map their domain to a common descriptor space. These approaches rely on a vast amount of annotated sketches.
Two relevant approaches, solving different tasks, is zero shot sketch-based image retrieval~\cite{drd+19} and deep shape matching~\cite{rtc18}. The latter trains a network for shape similarity on edge maps of landmark images, it has been, however, shown in ablations of~\cite{etc22} that training on rBTEs performs better in sketch recognition than training on plain edge maps.
\paragraph{Vision Transformers.}
Vision Transformers have recently been applied in visual recognition tasks, becoming a competitive architecture along Convolutional Neural Networks (CNNs).
A Vision Transformer splits an image in non-overlapping patches, which are called (visual) tokens.
Each patch is represented by an embedding vector, which is iteratively refined through a series of transformer-encoder layers~\cite{dbk+20}. 
DINO was proposed as a self-supervised pretraining method for Vision Transformers, that achieves impressive performance on downstream tasks~\cite{ctm+21}.
We use a DINO pretrained transformer to extract features in this work, as they have been shown to exhibit very nice properties, such as encoding powerful high-level information at high spatial resolution~\cite{agb+21}.

\section{Object Co-segmentation Method}
\label{sec:method}
In this section, we describe the necessary background and the proposed co-segmentation method in detail, as well as the novel application of it.

\subsection{Background} \label{sec:method-cl}
\paragraph{Deep Descriptor Transforming.}
In~\cite{wzw+19}, it is shown that deep descriptors coming from patches of images depicting a common object exhibit similar values. 
Given a collection of images of a common object, the vector representing the common object in the descriptor space is calculated as follows. 
Let $\{S_i\}_{i=1}^N$ be a set of descriptors of all patches collected from all images of the class, $\mu = \sum_{i=1}^N S_i / N$ be the mean of the descriptors, and $\mathbf{\xi}$ be
\begin{equation}
 \mathbf{\xi}^* = \arg \max_{\|\mathbf{\xi}\|=1} \sum_i (\mathbf{\xi}^\top (S_i-\mu))^2 \mbox{.}   \label{eqn:xi}
\end{equation}
Equation~(\ref{eqn:xi}) is efficiently solved by computing the eigenvector of $\sum (S_i-\mu)^\top (S_i-\mu)$ corresponding to the largest eigenvalue.
To estimate the common object relevance, a projection onto the direction $\mathbf{\xi}$ of each patch descriptor of each image is computed as
\begin{equation}
P_i = \xi^{*\top} (S_i - \mu).
\label{eqn:orig_proj}
\end{equation}

\noindent The common object relevance $R_i$ of patch $i$ within image $I$ is computed as
\begin{equation}
R_i = \frac{\max(0, P_i)}{\max_{j \in I}{P_j}}. \label{eqn:relevance}
\end{equation}
In other words, the negative responses of the background are set to zero, and the other responses are normalized into the interval $[0,1]$.
The result can be viewed as a common object probability map, given the set of images.

\paragraph{Normalized cut.}
Normalized cut (N-cut) is a spectral clustering method often used for image segmentation~\cite{shi+97}. 
Combining N-cut with DINO features was recently proposed in~\cite{wsh+22}.
For each image, a fully connected graph of all DINO features is represented by a two-valued affinity matrix $\mathcal{E}$ based on the cosine similarity
\begin{equation}
    \mathcal{E}_{i, j}= \begin{cases}1, & \text { if } \frac{v_{i}^\top v_{j}}{\left\|v_{i}\right\|_{2}\left\|v_{j}\right\|_{2}}\geq \tau \\ \epsilon, & \text { otherwise }\end{cases}
\end{equation}
where $v_i,v_j$ are Key embeddings of the patches $i,j$ respectively, $\tau$ is a similarity threshold and  $\epsilon$ is a small constant value to keep the graph fully connected.
Let $\mathcal{D}$ be a diagonal degree matrix with the diagonal elements $\mathcal{D}_i$ being a row-wise sum of $\mathcal{E}$, \ie $\mathcal{D}_i = \sum_{j}\mathcal{E}_{i, j}$.
Partitioning the image into foreground-background is performed by solving a generalized eigensystem 
\begin{equation}
(\mathcal{D}-\mathcal{E})y = \lambda\mathcal{D}y \mbox{,}     \label{eqn:geigsys}
\end{equation} 
in particular, by thresholding the eigenvector corresponding to the smallest non-zero eigenvalue~\cite{sma97}.

\paragraph{Biased N-cut.}
Biased N-cut~\cite{mvm11} is an extension of the N-cut algorithm, which can be guided by seed points in the image, pushing the foreground segmentation to include the seed points.
In particular, the segmentation is performed by thresholding the biased N-cut vector $\hat{x} = \sum_{k=2}^{K+1} w_k u_{k} \mbox{,} $
which is a linear combination of $K$ eigenvectors which correspond to the $K$ smallest non-zero eigenvalues of the generalized eigensystem~(\ref{eqn:geigsys}).
The weights are
\begin{equation}
w_k = \frac{u_{k}^{\top} (\mathcal{D} s)}{\lambda_{k}-\gamma},
\end{equation}
where $\mathcal{D}$ is the degree matrix and $s$ is the seed weight vector indicator, 
where each dimension $s_i$ corresponds to the weight of $i$-th pixel (or patch in our case),
$\lambda_k$ is the $k$-th eigenvalue and $\gamma$ is a hyperparameter.

\subsection{Class-Biased Normalized-Cut} 
\begin{figure}[t]
\begin{center}
    \includegraphics[trim={0cm 7.3cm 0cm 3cm} ,width=1.0\textwidth]{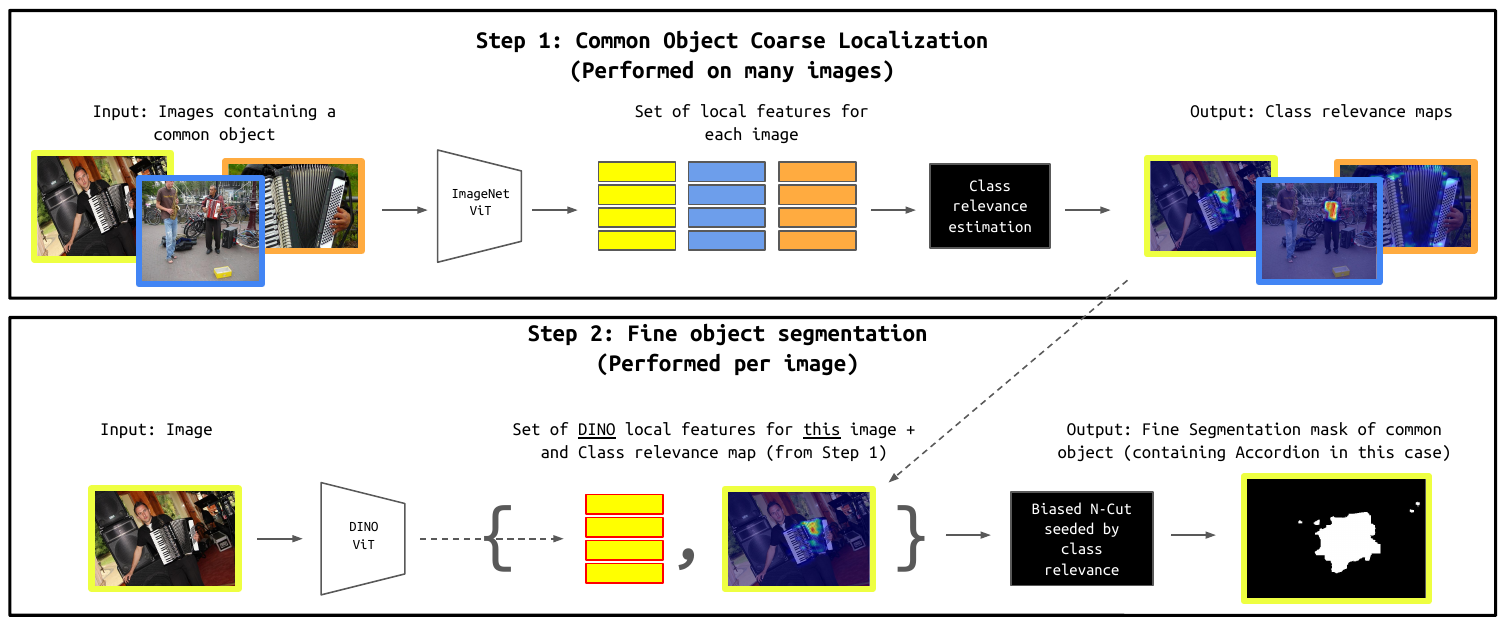}
\end{center}
\caption{
A diagram presenting the pipeline of the proposed co-segmentation method. This pipeline is also used as a preprocessing for the training set of the sketch recognition task.
}
\label{fig:pipeline}
\end{figure}

The proposed method, named Class-Biased Normalized-Cut (CBNC) takes a set of RGB images that contain a common object (a class) as input and outputs a set of binary co-segmentation masks. 
It consists of the following steps, which are also presented in Fig.~\ref{fig:pipeline} visually.

\paragraph{Common Object Coarse Localization.}
For each class in turn, patch features are extracted for all example images of that class. 
An ImageNet pre-trained  Vision Transformer~\cite{dbk+20} is used to extract the patch features. Each image is split into $(h\times w)$ non-overlapping patches of size $K\times K$ (square), where $(h,w)$ are given by $h = H / K, w = W / K$; $H,W$ being the height and the width of the image in pixels.
The Keys of the last self-attention layer of the ViT are used as the patch descriptors.
We modify eqn.~(\ref{eqn:orig_proj}) to be:
\begin{equation}
P_i = \sigma \xi^{*\top} (S_i - \mu),
\end{equation}
where $\sigma \in \{-1,1\}$ is a sign that fixes the foreground / background ambiguity -- note that both $\mathbf{\xi}$ and $-\mathbf{\xi}$ are solutions to eqn.~(\ref{eqn:xi}). The sign $\sigma$ is selected so that the majority of image-border patches across all the class images is $P_i < 0$; that is, the majority (not necessarily all and in all images) of the image-border patches belong to the background.
We calculate $P_i$ and subsequently $R_i$
for each patch in the class, forming the class relevance heatmaps.

\begin{figure}

\parbox{.47\linewidth}{
\centering
 ``hat'' class \\
 \scalebox{0.7}{
\begin{tabular}{@{\ssp}c@{\ssp}c}
\includegraphics[height=2.3cm, width=2.3cm]{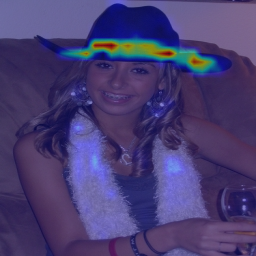}
&
\includegraphics[height=2.3cm, width=2.3cm]{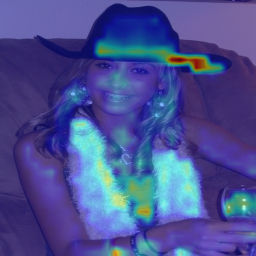}
\end{tabular}
}
\\
``avocado'' class\\
\scalebox{0.7}{
\begin{tabular}{@{\ssp}c@{\ssp}c}
\includegraphics[height=2.3cm, width=2.3cm]{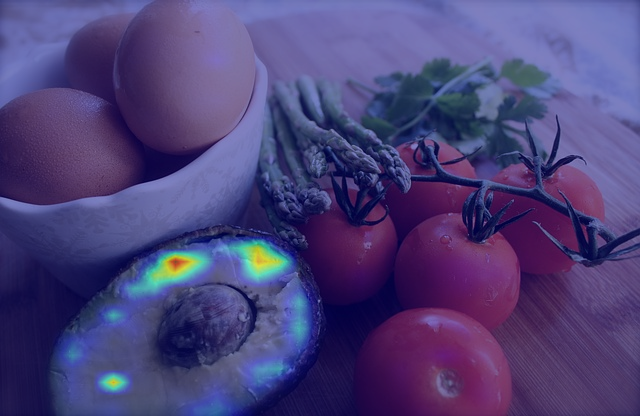}
&
\includegraphics[height=2.3cm, width=2.3cm]{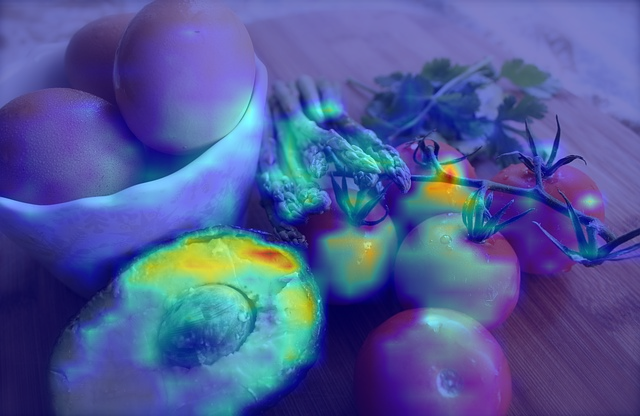}
\end{tabular}
}

\caption{
Comparison of class relevance heatmaps produced by ImageNet-ViT (left) vs. DINO-ViT (right) for two different classes. 
It is observed that ImageNet features provide more discriminative class relevance.
}
\label{fig:dino_vs_imagenet_heatmap}
}
\hfill
\parbox{.47\linewidth}{        \centering
    \scalebox{0.65}{
    \begin{tabular}{@{}c@{\msp}c@{\msp}c@{}}
    Class relevance $R_i$ & Eroded & Soft-max bias $s_i$\\
         \includegraphics[height=2.5cm, width=2.5cm]{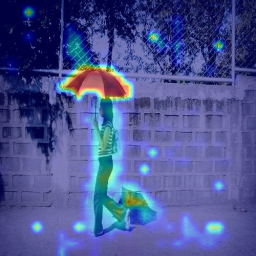}&
         \includegraphics[height=2.5cm, width=2.5cm]{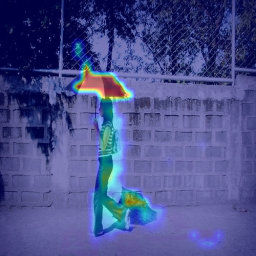}&
         \includegraphics[height=2.5cm, width=2.5cm]{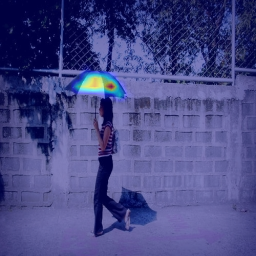}\\
    \end{tabular}
    }
\caption{Visualization of the class relevance eqn.~(\ref{eqn:relevance}) (left), eroded class relevance (middle), and soft-max bias of the eroded class relevance (right) used as seed weights to bias the N-Cut algorithm, for a sample of the ``umbrella'' class. The importance of the two further refinements of the class relevance heatmap is observed.}
\label{fig:relevance}
}
\end{figure}

In~\cite{agb+21}, DINO features were used for class relevance.
However, while DINO features capture well part consistency of objects~\cite{agb+21}, we choose 
ImageNet trained ViT features for this step.
The ImageNet ViT model is trained with a classification objective, and hence, similar token responses are
generated across various class instances.
The strongest class
relevance responses typically focus on common salient
parts.
This is confirmed by our experiments: a qualitative difference is shown in Figure~\ref{fig:dino_vs_imagenet_heatmap}, a quantitative comparison is provided in Table~\ref{tab:ablation2}. 
Nevertheless, DINO is trained using relatively small cut-outs,
therefore, it provides highly similar intra-image responses
on the object, which is suitable for segmentation.
These abilities of DINO features are exploited in the next step.

In~\cite{wzw+19}, the class relevance was used directly as localization masks by considering all the patches that have positive relevance as foreground.
However, we observe that this approach almost always delivers regions covering more than the common object of the class due to, for example, the appearance of another common object in many of the class images, e.g., umbrella-person (see Figure~\ref{fig:relevance} (left)).
This is why we use the class relevance heatmaps only as a first indicator of the common object and not the final mask, which we calculate in the next step.

\begin{figure}[h]
\centering
\scalebox{0.75}{
\begin{tabular}{@{\ssp}c@{\ssp}c@{\ssp}c@{\ssp}c}
\includegraphics[height=2.2cm, width=2.2cm]{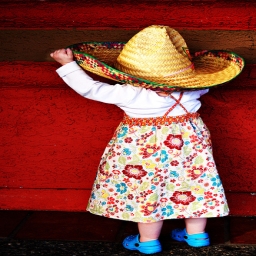}
&
\includegraphics[height=2.2cm, width=2.2cm]{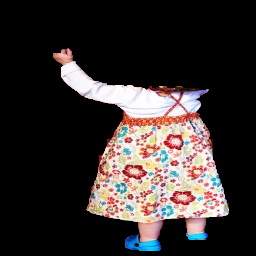}
&
\includegraphics[height=2.2cm, width=2.2cm]{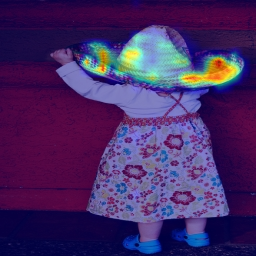}
&
\includegraphics[height=2.2cm, width=2.2cm]{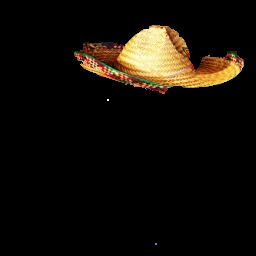}
\\
(a) & (b) & (c) & (d) \\
\end{tabular}
}
\caption{
(a) Image of the class ``hat''.
(b) Segmentation obtained by the N-cut method.
(c) Class relevance heatmap, values of 0 are in deep blue, positive values range from blue (low) to red (high).
(d) Segmentation obtained by the proposed method, using the class relevance bias.
Without the class information, the N-cut segmentation fails to focus on the class object.}
\label{fig:ncut_fail}
\end{figure}

\begin{figure}[ht]
    \centering
    \scalebox{0.7}{
    \begin{tabular}{@{}ccc@{}}
         \includegraphics[height=2.45cm, width=2.45cm]{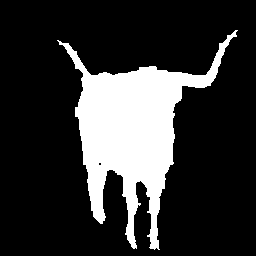}&
         \includegraphics[height=2.45cm, width=2.45cm]{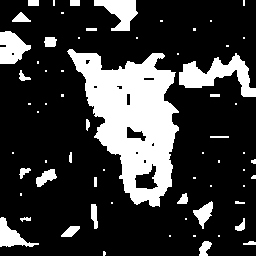} &
         \includegraphics[height=2.45cm, width=2.45cm]{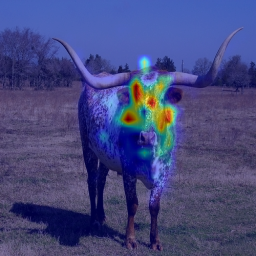}\\
    \end{tabular}
    }
\caption{Qualitative comparison of the biased N-Cut segmentation using affinity from DINO ViT features (left) and ImageNet ViT features (middle). Both use the class-relevance bias estimated by ImageNet ViT features (right).}
\label{fig:imagenet_affin}
\end{figure}

\paragraph{Fine Object Segmentation.}
This step is split into two subprocedures, namely (1) Biased N-cut seeded by class relevance and (2) Fine Boundary refinement.
\\

\noindent \textbf{Biased N-cut seeded by class relevance.}
Since N-cut operates on every image individually, it is not guaranteed that the cluster corresponds to the common object of a set of images, see Figure~\ref{fig:ncut_fail}.
In the proposed method, we combine the intra-image similarity used by the N-cut with the information collected from all class examples in the form of a seed bias. 
In particular, we use the soft biased N-cut~\cite{mvm11} guided by the class relevance to bias the segmentation towards the common object of the set of images.
The seed vector $s$ is directly proportional to the class relevance heatmap.
In particular, the non-zero relevance is first eroded by a $2 \times 2$ mask to remove small random responses and border responses that partially include the background. 
In order to promote confident areas, a soft-max function with temperature $\beta$ is applied to the remaining elements of the object relevance map to obtain weights $s_i$ for each image patch.
The class relevance, its eroded version and the final class-relevance bias are visualized in Figure~\ref{fig:relevance}.
We observe that DINO ViT features are crucial for this step (to build the affinity matrix $\mathcal{E}$) and that ImageNet-trained ViT features perform poorly; see Figure~\ref{fig:imagenet_affin} for qualitative examples. A quantitative comparison is provided in Table~\ref{tab:ablation2}.
\\

\noindent \textbf{Fine boundary refinement.}
The previous steps operate on ViT tokens which correspond to $8 \times 8$ image patches. 
The coarse segmentation mask is finally upscaled and refined by standard GrabCut~\cite{rkb04} implementation~\cite{b00}.

\subsection{Application: Sketch Classification}
\label{sec:skclass}
We apply the proposed co-segmentation method as a preprocessing step to remove the background clutter from the training images, as our assumption is that this will improve recognition performance on real sketches.
Prior to classifier training, images of each class are processed to localize the object of interest in them. Similarly to~\cite{etc22}, rBTEs are extracted, but only from the area predicted to be occupied by the object. The classifier is then trained using rBTEs, as in~\cite{etc22}.

\begin{figure}[ht]
\centering
\scalebox{0.72}{
\begin{tabular}{@{\ssp}c@{\ssp}c@{\ssp}c@{\ssp}c}
``snake'' & ``camel'' & ``airplane'' & ``eyeglasses''\\
\begin{tabular}{@{\ssp}c@{\ssp}c}
\includegraphics[height=1.9cm, width=1.9cm]{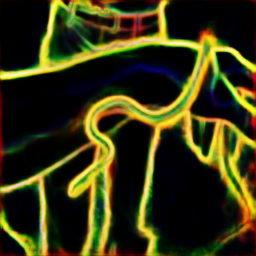}
&
\includegraphics[height=1.9cm, width=1.9cm]{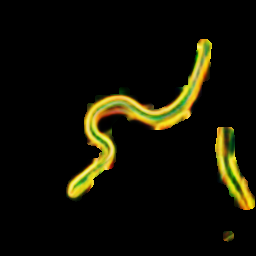}\\
\includegraphics[height=1.9cm, width=1.9cm]{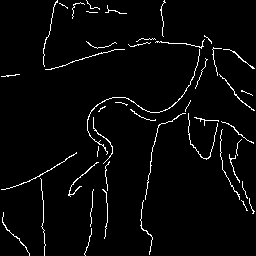}&
\includegraphics[height=1.9cm, width=1.9cm]{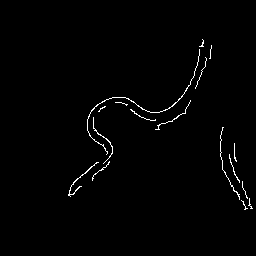}\\
\end{tabular}
&
\begin{tabular}{@{\ssp}c@{\ssp}c}
\includegraphics[height=1.9cm, width=1.9cm]{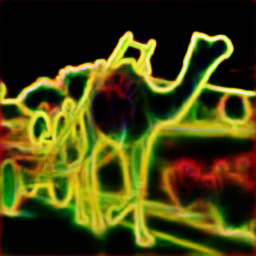}
&
\includegraphics[height=1.9cm, width=1.9cm]{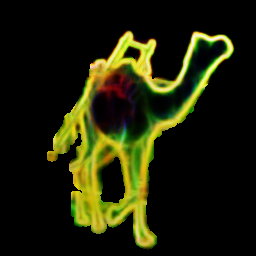}
\\
\includegraphics[height=1.9cm, width=1.9cm]{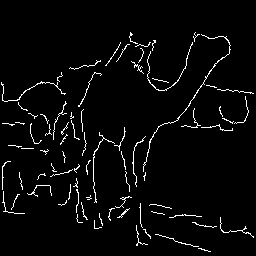}
&
\includegraphics[height=1.9cm, width=1.9cm]{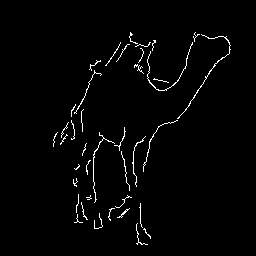}
\\
\end{tabular}
&
\begin{tabular}{@{\ssp}c@{\ssp}c}
\includegraphics[height=1.9cm, width=1.9cm]{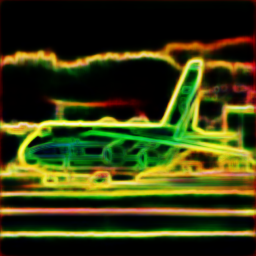}
&
\includegraphics[height=1.9cm, width=1.9cm]{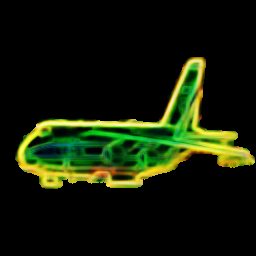}
\\
\includegraphics[height=1.9cm, width=1.9cm]{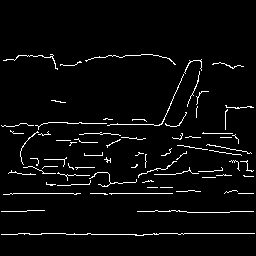}
&
\includegraphics[height=1.9cm, width=1.9cm]{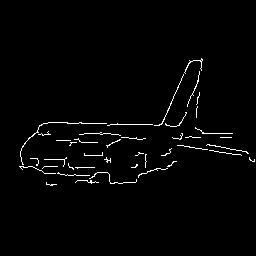}
\\
\end{tabular}
&
\begin{tabular}{@{\ssp}c@{\ssp}c}
\includegraphics[height=1.9cm, width=1.9cm]{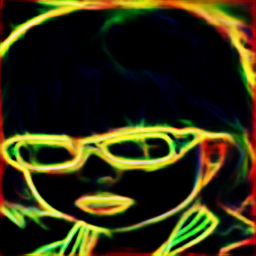}
&
\includegraphics[height=1.9cm, width=1.9cm]{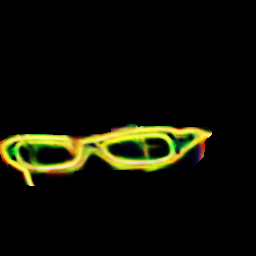}
\\
\includegraphics[height=1.9cm, width=1.9cm]{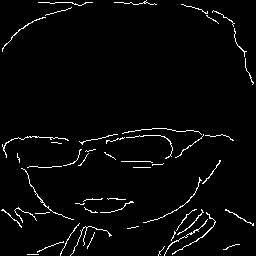}
&
\includegraphics[height=1.9cm, width=1.9cm]{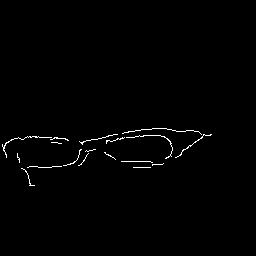}
\end{tabular}
\end{tabular}
}
\caption{
Examples from four classes of the Sketchy dataset.
For each, we show on the top row an edgemap produced by the baseline method~\cite{etc22} (left) and by our method (right), and their corresponding rBTEs on the bottom row.
}
\label{fig:results}
\end{figure}

The edgemap for each image is formed following the baseline \cite{etc22}, and one of the following detectors Structured Forests \cite{dz13}, HED \cite{sz17} and BCDN \cite{jsm+19} is randomly used to extract the edges. For one example image, different detectors can be used in different epochs. The edges are masked with the binary segmentation mask, and edges outside the mask are dropped.
Such an edge map is finally processed by the rBTE pipeline as proposed in~\cite{etc22} to form the training examples.
In Figure \ref{fig:results}, examples of the edge maps fed into the rBTE pipeline are shown for both approaches, \cite{etc22} and ours together with their corresponding rBTEs.  All three detectors are visualized at the same time, one per RGB channel.

\section{Experiments and Results}
\label{sec:experiments}
The proposed method (CBNC) is experimentally evaluated directly on the co-segmentation task and on the application task of sketch classification.
Implementation details can be found in the Appendix.

\begin{table}[t]
  \centering
    \caption{Comparison of co-segmentation on the PASCAL VOC dataset. 
  Two standard measures are reported: the Mean Jaccard Index ($J_m$) -- higher is better, and Precision ($P_m$) - higher is better. 
  The top-performing approach is highlighted in bold font.
  The proposed method surpasses all other methods that use the same level of supervision (image-level) while closing the performance gap with ones that use costly pixel-level supervision.
  }
  \newcommand{\bl}[1]{\color{blue}{#1}}
\scalebox{0.89}{
\begin{tabular}{cc|cc}
\hline
Method  & Supervision & $J_m \uparrow$ & $P_m \uparrow$  \\
\hline
SSNM~\cite{zcl+20} & Pixel & 71.0 & 94.9 \\
DOCS~\cite{lhr19} & Pixel & 65.0 & 94.2 \\
CycleSegNet~\cite{zll+21} & Pixel & { 75.4} &{ 95.8} \\
Li et al.~\cite{lsl+19} & Pixel & 63.0 & 94.1 \\
\hline
N-cut~\cite{wsh+22} & Image & 57.8 & 90.6 \\
GO-FMR~\cite{qhz+16} & Image & 52.0 & 89.0 \\
Hsu et al.~\cite{hlc+18} & Image & 60.0 & 91.0 \\
Amir et al.~\cite{agb+21} & Image & 60.7 & 88.2 \\
\hdashline
\method (ours) & Image & {\bf 64.7 } & {\bf 92.0} \\
\hline
\end{tabular}
}

  \label{tab:real_coseg}
\end{table}

\begin{table*}[t]
  \centering
    \caption{Comparison of co-saliency detection results on three challenging datasets. Three standard measures are reported: the Mean Average Error ($MAE$) -- lower is better, the maximum of F-measure ($F_{\beta}^{\max}$) -- higher is better, and the S-measure ($S_{\alpha}$) - higher is better.
  The top-performing approach is highlighted in bold font.
  The proposed method surpasses all other methods that use the same level of supervision (image-level) while closing the performance gap with ones that use pixel-level costly supervision.
  \label{tab:cosal}
  }
  \newcommand{\bl}[1]{\color{blue}{#1}}

\scalebox{0.86}{
\begin{tabular}{cc|ccc|ccc|ccc}
\hline
& &\multicolumn{3}{|c}{CoSal2015~\cite{zhl+16}}& \multicolumn{3}{|c}{CoSOD3k~\cite{flj20}}& \multicolumn{3}{|c}{CoCA~\cite{zjx+20}}\\ 
\hline
Method  & Supervision & $MAE \downarrow$ & $ F_{\beta}^{\max}\uparrow$ &$S_{\alpha}\uparrow$ & $MAE \downarrow$ & $ F_{\beta}^{\max}\uparrow$ &$S_{\alpha}\uparrow$ & $MAE \downarrow$ & $ F_{\beta}^{\max}\uparrow$ &$S_{\alpha}\uparrow$ \\
\hline
CADC~\cite{zhl+21} & Pixel & {0.064} & { 0.862} & { 0.866} & 0.096 & 0.759 & 0.801 & 0.132 & 0.548 & 0.681\\
UFO~\cite{sds+22} & Pixel & {0.064} & { 0.865} & {0.860} & 0.073 & {0.797} & {0.819} & {0.095} & {0.571} & {0.697 }\\
GCoNet~\cite{fff+21} & Pixel & 0.068 & 0.847 & 0.845 &  0.071 & 0.777 & 0.802 & 0.105 & 0.544 & 0.673 \\
DCFM~\cite{yxz+22} & Pixel & 0.067 & 0.856 & 0.838 & {0.067} & {0.805} & 0.810 & { 0.085} & { 0.598} & { 0.710} \\
\hline
Li et al.~\cite{lsw+19} &  Image & - & 0.712 & 0.763   & - & - & - & - & - & - \\
UCSG~\cite{htl+18} &  Image & - & 0.758 & 0.751 & - & - & - & - & - & - \\
Amir et al.~\cite{agb+21} & Image & 0.100 & 0.770 & 0.796 & 0.119 & 0.701 & 0.753 & 0.253 & 0.389 & 0.549\\
N-cut~\cite{wsh+22} & Image & 0.087 & 0.811 & 0.818 & 0.092 & 0.756 & 0.786 & 0.143 & 0.481 & 0.637\\
\hdashline
\method (ours) & Image & { \bf  {0.056} } &  { \bf 0.857 }& {\bf 0.856} & {\bf 0.068} & {\bf 0.792} & {\bf 0.815} & {\bf 0.125} & {\bf 0.528} & {\bf 0.668}\\
\hline
\end{tabular}
}

\end{table*}

\begin{figure*}[ht]
\scalebox{0.82}{

\centering
\begin{tabular}{cccccccc}
Class Name & Sample RGB & Ground Truth & Ours & UFO & GCoNet & DCFM &
Amir et al.\\
\hline \\
\raisebox{0.75cm}{violin} &
\includegraphics[height=1.6cm, width=1.6cm]{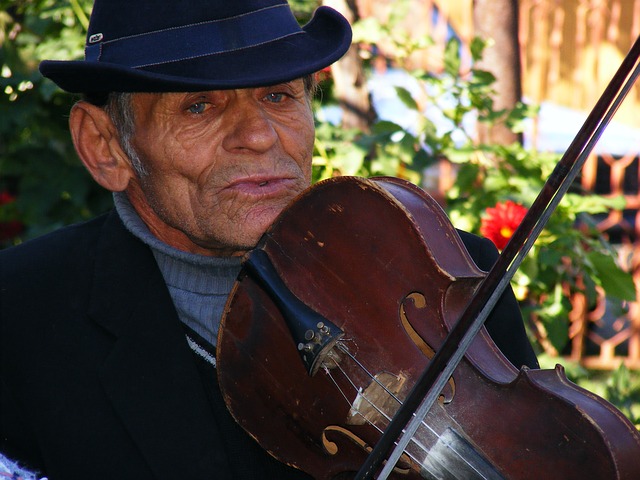}&
\includegraphics[height=1.6cm, width=1.6cm]{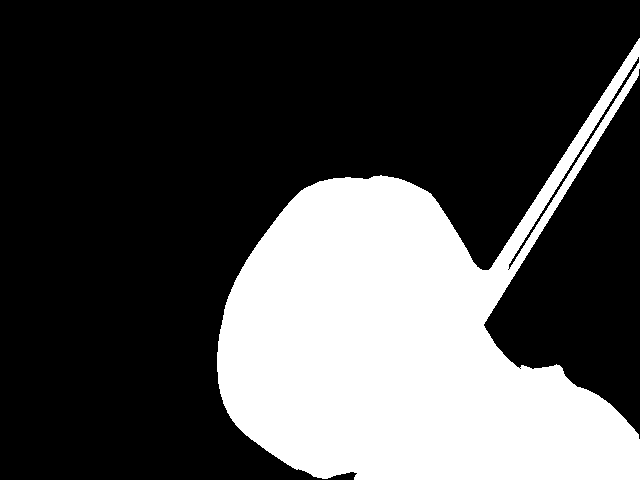}&
\includegraphics[height=1.6cm, width=1.6cm]{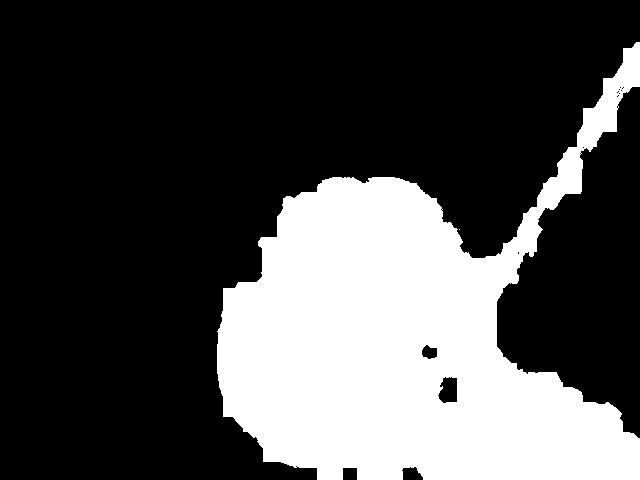}&
\includegraphics[height=1.6cm, width=1.6cm]{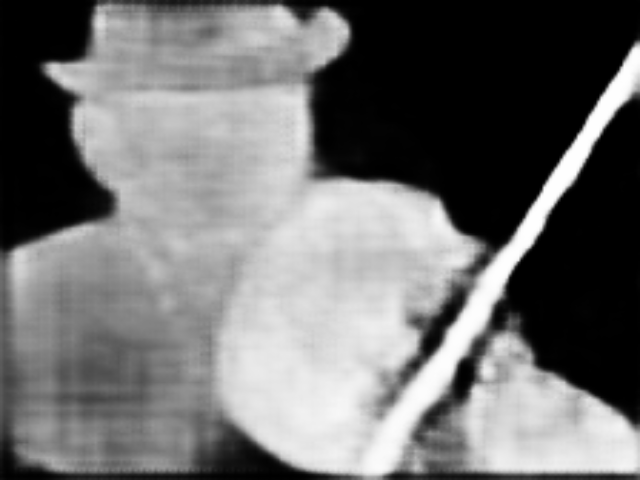}&
\includegraphics[height=1.6cm, width=1.6cm]{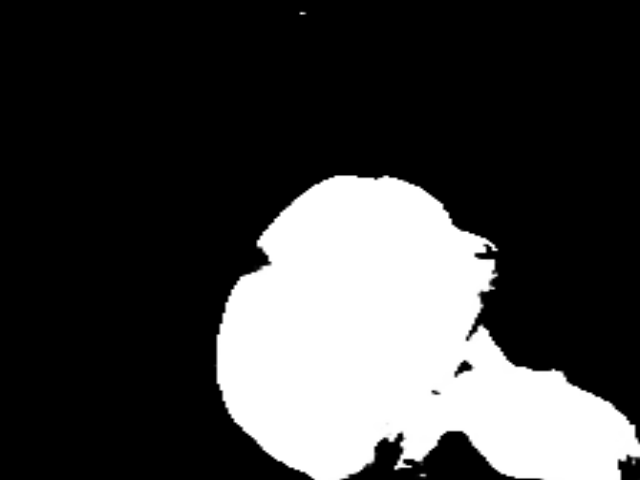}&
\includegraphics[height=1.6cm, width=1.6cm]{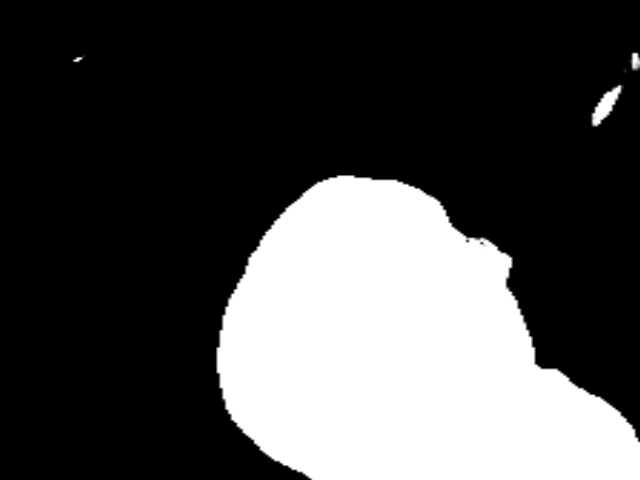}&
\includegraphics[height=1.6cm, width=1.6cm]{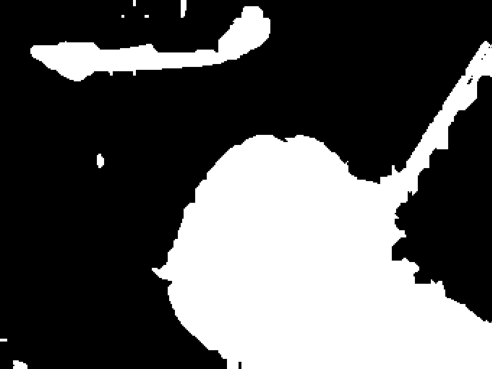}\\
\raisebox{0.75cm}{pineapple}&
\includegraphics[height=1.6cm, width=1.6cm]{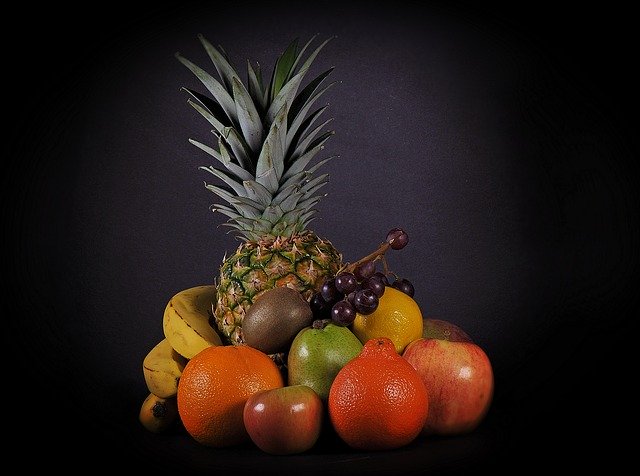}&
\includegraphics[height=1.6cm, width=1.6cm]{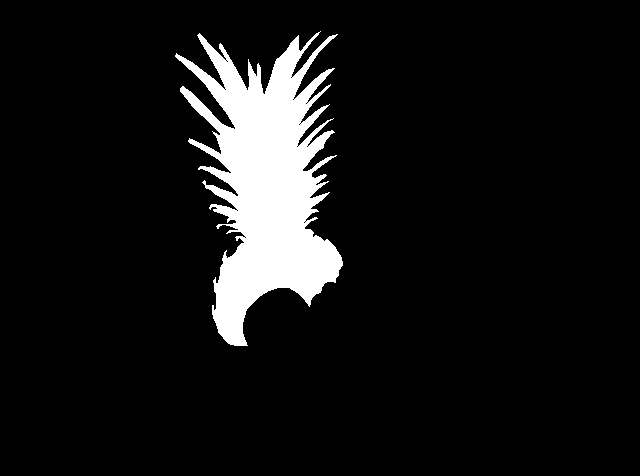}&
\includegraphics[height=1.6cm, width=1.6cm]{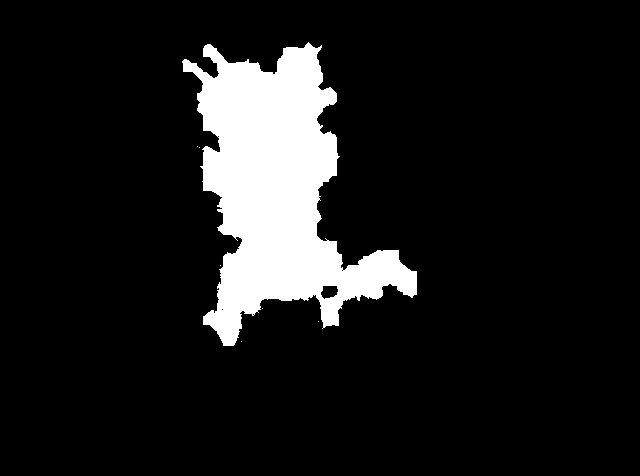}&
\includegraphics[height=1.6cm, width=1.6cm]{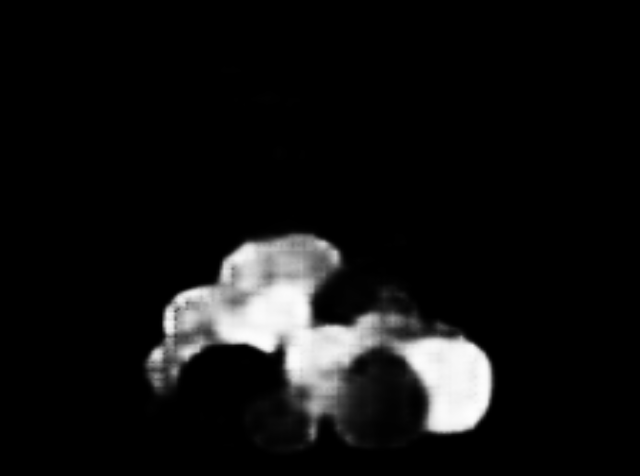}&
\includegraphics[height=1.6cm, width=1.6cm]{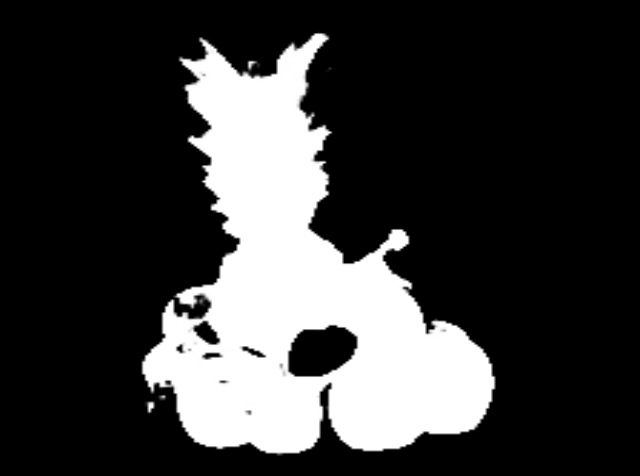}&
\includegraphics[height=1.6cm, width=1.6cm]{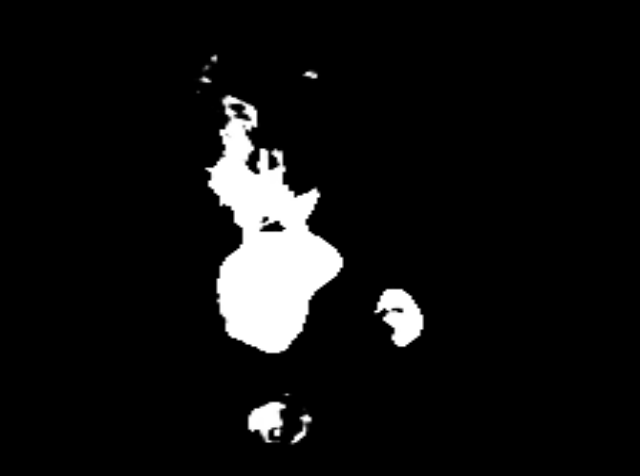}&
\includegraphics[height=1.6cm, width=1.6cm]{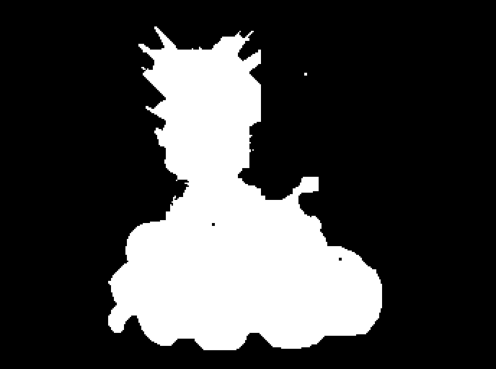}\\
\raisebox{0.75cm}{guitar}&
\includegraphics[height=1.6cm, width=1.6cm]{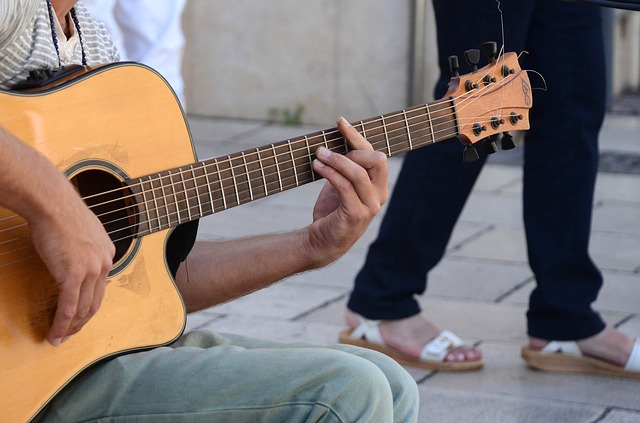}&
\includegraphics[height=1.6cm, width=1.6cm]{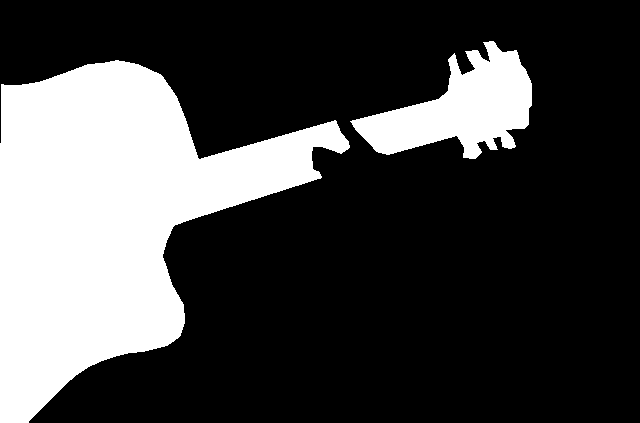}&
\includegraphics[height=1.6cm, width=1.6cm]{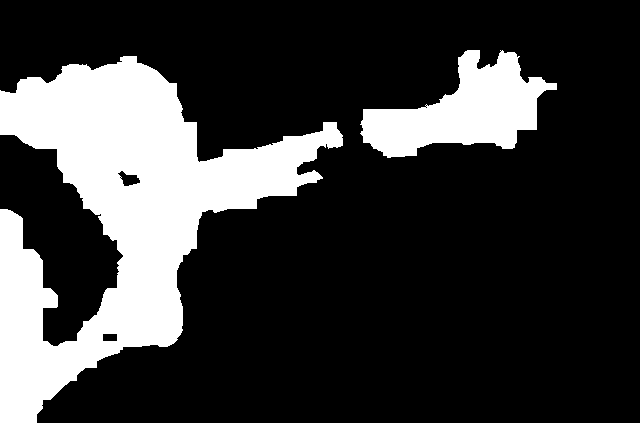}&
\includegraphics[height=1.6cm, width=1.6cm]{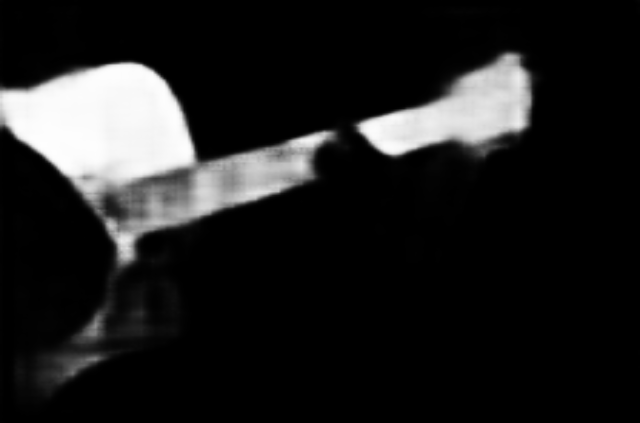}&
\includegraphics[height=1.6cm, width=1.6cm]{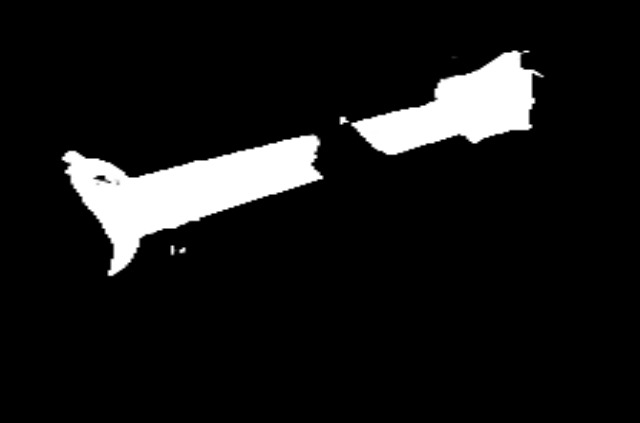}&
\includegraphics[height=1.6cm, width=1.6cm]{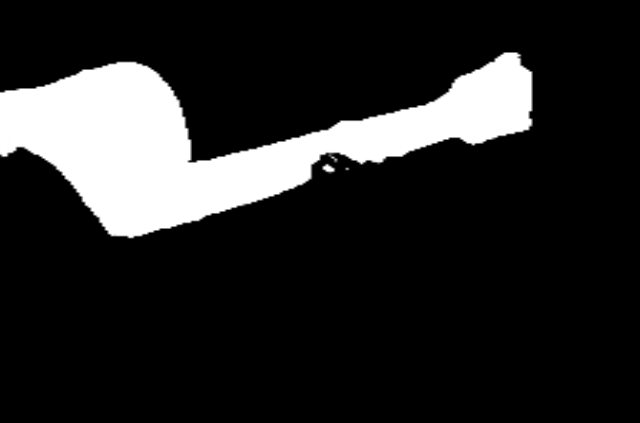}&
\includegraphics[height=1.6cm, width=1.6cm]{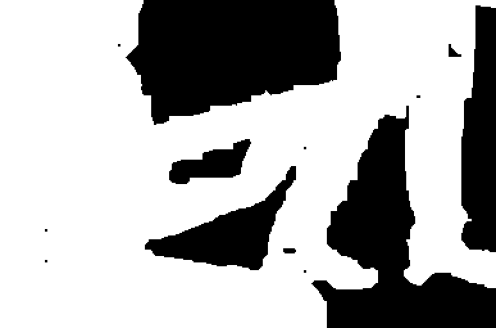}\\
\end{tabular}
}
\caption{Co-saliency detection results for cases in which our method succeeds, but others fail. 
Each row corresponds to one sample coming from the class mentioned in the first column; the classes come from any of the CoSal2015~\cite{zhl+16}, CoSOD3k~\cite{flj20} and CoCA~\cite{zjx+20} datasets.
The ground truth segmentation mask is shown for reference.
}
\label{fig:results_cosal}
\end{figure*}

\subsection{Co-Segmentation and Co-Saliency detection}
Firstly, we evaluate the proposed method on the most challenging co-segmentation dataset, the PASCAL VOC dataset~\cite{evw+10}, as used in~\cite{fi13}.
Additionally, we evaluate the proposed method on 3 challenging co-salient object detection datasets, namely CoSal2015~\cite{zhl+16}, CoSOD3k~\cite{flj20}, and CoCA~\cite{zjx+20}.
Both tasks of co-segmentation and co-salient object detection offer ground truth in the form of binary masks.
In the former, methods should produce a binary prediction. However, this constraint is lifted in the latter, and the model predicts a grayscale image in the range $[0,1]$.
In both cases, the input to a method is just the set of images.
For co-segmentation, the evaluation metrics used include Jaccard Index ($J_m$) and Precision ($P_m$), while for co-salient object detection the Mean Absolute Error ($MAE$)~\cite{cwl+13}, maximum F-measure ($F_{\beta}^{\max}$)~\cite{ahe+09} and S-measure ($S_{\alpha}$)~\cite{fcl+17} are calculated.
We compare with recent state-of-the-art approaches that use the same level of supervision that we use to train (denoted as ``Image'' level supervision).
For reference, we include methods that use pixel-level supervision in the form of ground truth segmentation masks.

The results are presented in Tables \ref{tab:real_coseg} and \ref{tab:cosal}, indicating that our method outperforms all other methods that use the same level of supervision (image-level) while approaching or being on par with methods that use expensive pixel-level supervision. 
Interestingly, simple N-Cut~\cite{wsh+22} performs competitively despite working with each image of the set separately.
We present examples of our segmentations and ones from state-of-the-art methods in Figure~\ref{fig:results_cosal}.

\paragraph{Ablation.}
We compare the results of co-salient detection on the CoCA dataset by using either DINO or ImageNet classification ViT features for each of the steps of the proposed method. The results are presented in Table \ref{tab:ablation2}.
Quantitative results agree with qualitative ones, which indicate that ImageNet ViT features on the N-Cut step are not suitable, see Figure~\ref{fig:imagenet_affin}, and that DINO ViT features are not suitable for the class relevance step.
A qualitative comparison of the class relevance estimates is provided in Figure~\ref{fig:dino_vs_imagenet_heatmap}.

\begin{table}\parbox{.47\linewidth}{
\centering    
\caption{
  Co-saliency detection results on the CoCA dataset. Comparison of the use of DINO ViT features vs. ImageNet ViT features for each of the two steps of the method.
  The proposed combination of ImageNet features for inter-image class token relevance, and DINO features for intra-image token relevance provides the best performance.
  \label{tab:ablation2}
  }
\scalebox{0.8}{
\begin{tabular}{@{\msp}l@{\msp}cccc}
\toprule
Class relev. & Intra-image & $MAE \downarrow$ &$ F_{\beta}^{\max}\uparrow$ &$S_{\alpha}\uparrow$  \\
\midrule
ImageNet & ImageNet & 0.268 & 0.321 & 0.483 \\ 
DINO & DINO & 0.161 & 0.465 & 0.626 \\ 
ImageNet & DINO & \textbf{0.125} & \textbf{0.528} & \textbf{0.668} \\ 
\bottomrule
\end{tabular}
}
} 
\hfill
\parbox{.47\linewidth}{
\caption{
Co-saliency detection results on the CoCA dataset. 
Study of the effect of the presence of non-class images on the final co-segmentation metrics.
The robustness of the proposed method to outlier images in the class-relevance step is observed.
\label{tab:distractors}
}
\centering    
\scalebox{0.8}{
\begin{tabular}{@{\msp}c@{\msp}ccc}
\toprule
\# Non-class images & $MAE \downarrow$ &$ F_{\beta}^{\max}\uparrow$ &$S_{\alpha}\uparrow$  \\
\midrule
0 & 0.125 & 0.528 & 0.668 \\ 
10 & 0.130 & 0.517 & 0.660 \\ 
20 & 0.135 & 0.518 & 0.657 \\ 
\bottomrule
\end{tabular}
}
}
\end{table}

\paragraph{Robustness to outliers.}
The robustness of the class relevance estimation step to the presence of outlier images that do not contain the class object was evaluated on the CoCA dataset.
The model was learned with up to 20 additional images sampled from other categories at random; the segmentation quality was evaluated in a standard way.
The results shown in Table~\ref{tab:distractors} show that even with 20 outlier images (the average number of class examples is $16.2$), no significant drop in performance is observed.

\subsection{Sketch classification}
We perform experiments on the Sketchy dataset~\cite{sbh+16} to evaluate the improvement of the baseline method trained with our preprocessed training data over the baseline~\cite{etc22} and to compare
different co-segmentation methods on this task. 
The Sketchy dataset was initially constructed for the task of sketch-based image retrieval, and it consists of sketches and natural images from 125 categories.
We follow the training and evaluation protocol of~\cite{etc22}, based on which the training set consists of 11,250 photographs, while the evaluation set consists of 7,063 sketches.
We use the same hyperparameters for both the baseline and our method, with the only difference being the examples used for training the deep classifier (original and transformed by our method), as we want to evaluate the impact of the transformed training set.
We refer to the baseline method as ``Baseline rBTE'' and to our method as ``CBNC rBTE''.
We report Top-1 and Top-5 balanced classification accuracy (\%).
We report results using the ImageNet pre-trained ResNet18 (R18) and ResNet101 (R101) backbones in Table \ref{tab:Sketchy}. 
We observed a large improvement when using the training examples produced by our method for both backbones.

\begin{table}\parbox{.47\linewidth}{
  \centering
\caption{Comparison of ResNet18 and ResNet101 performance on the Sketchy dataset.
  The baseline method trained on original edgemaps and the proposed method trained on segmented edgemaps are compared.
  \label{tab:Sketchy}
  }
  \scalebox{0.9}{
  \scalebox{0.9}{
\begin{tabular}{lccc}
\toprule
Method     & Backbone     & Top-1     & Top-5 \\
\midrule
Baseline rBTE~\cite{etc22}     & R18 & 40.9  & 67.4  \\
\midrule
\method rBTE (ours)  & R18 & \textbf{50.0} & \textbf{77.1} \\
\midrule
\midrule
Baseline rBTE~\cite{etc22}  & R101 & 49.7 &    74.1  \\
\midrule
\method rBTE (ours) & R101 & \textbf{56.6} & \textbf{80.6} \\
\bottomrule
\end{tabular}
}
  }

} 
\hfill
\parbox{.47\linewidth}{   
  \centering
  \caption{ Comparison of different segmentation methods on Sketchy dataset. The "class info" column indicates whether the segmentation operates on individual images, or whether it uses information collected over all class images. The classification backbone used for this experiment is ResNet18.
  }
  \scalebox{0.9}{
  \scalebox{0.92}{
\begin{tabular}{@{\ssp}l@{\ssp}ccc}
\toprule
Segmentation     & Class info & Top-1     & Top-5 \\
\midrule
None (Baseline) \cite{etc22}   & \xmark & 40.9  & 67.4     \\
Class relevance~\cite{wzw+19}   & \cmark & 43.7 & 70.7 \\
N-cut~\cite{wsh+22} (no bias)  & \xmark & 47.9 & 74.7 \\
Amir et al~\cite{agb+21}   & \cmark & 43.1  & 68.4 \\
\midrule
\method rBTE (ours)   & \cmark & \textbf{50.0} & \textbf{77.1} \\ 
\bottomrule
\end{tabular}
}
  }

  \label{tab:ablation}
}
\end{table}

\paragraph{Segmentation methods.}
We evaluate different segmentation methods, ranging from no segmentation, which is the baseline~\cite{etc22}, using positive class relevance~~\cite{wzw+19}, N-cut based method without the class-relevance bias, the recent method of Amir et al.~\cite{agb+21}, and our proposed method based on biased N-cut. In all methods, ImageNet ViT features are used for class relevance, and the DINO ViT features for N-Cut affinity matrix construction are used. 
The results in Table~\ref{tab:ablation} show that any of the segmentation methods outperform the baseline. Hence, the assumption of the background edge clutter downgrading the recognition performance is verified. 
The best performance is achieved by the proposed method using the biased N-cut. 

\paragraph{Large-scale results.}
We perform an experiment on the large-scale Im4Sketch dataset, which was recently created for the task of photograph-to-sketch recognition~\cite{etc22}.
It contains 1,007,878 photographs for training from 874 classes and an evaluation set of 80,582 sketches from 393 classes, which come from a number of sketch classification datasets.
Since there are more classes in the training set than in the sketch evaluation dataset, two evaluation protocols are proposed in~\cite{etc22}. The classifier is always trained with all training classes. In evaluation, the classification is performed either for all classes or only for the set of test classes.
In this experiment, the backbone is the ImageNet pretrained ResNet101. 
We also follow the training and evaluation protocol of~\cite{etc22} to compare with the baseline.
The results are summarized in Table~\ref{tab:im4sketch}, verifying the effectiveness of the proposed preprocessing step as a means to improve sketch recognition performance. 

\begin{table}[ht]
 \centering
 \caption{Results for the Im4Sketch dataset. Softmax full: classification into all training classes is evaluated, subset: only classes in the test set are considered. 
}
 \scalebox{0.93}{
\begin{tabular}{l@{\ssp}ccc@{\ssp}}
\toprule
Method    & Softmax & Top-1 & Top-5 \\
\midrule
Baseline \cite{etc22} & full & 11.3  & 22.2     \\
\midrule
\method rBTE (ours) & full & \textbf{13.3} & \textbf{25.5} \\
\midrule
\midrule
Baseline \cite{etc22} & subset& 12.7 & 25.3     \\
\midrule
\method rBTE (ours) & subset & \textbf{17.6} & \textbf{32.7} \\
\bottomrule
\end{tabular}
}
\label{tab:im4sketch}
\end{table}

\section{Conclusions}
We proposed a novel method for object co-segmentation that uses a combination of two pre-trained Vision Transformer (ViT) models, exploiting the strengths of each while using no expensive pixel-level annotations for training.
The effectiveness of the approach was verified through experiments on challenging benchmarks, achieving the state-of-the-art among methods that use the same level of supervision while approaching or even surpassing methods that use more expensive pixel-level supervision.

The benefits of the co-segmentation method were further demonstrated on large-scale sketch classification.
The approach of learning the classifier on a proxy domain derived from natural images, without a single sketch involved in training, was adopted by using the rBTE domain~\cite{etc22}.
Unlike prior work, we proposed to suppress the edges of background clutter by estimating the spatial extent of the object using the proposed co-segmentation pipeline, significantly boosting classification performance.

\noindent \textbf{Limitations.}
One limitation of the proposed co-segmentation method is that the class relevance estimation step sometimes fails on image sets that are consecutive video frames - both the object and the background are common content in the set of images.

\newpage


\bibliographystyle{splncs04}
\bibliography{bib}

\begin{thebibliography}{10}
\providecommand{\url}[1]{\texttt{#1}}
\providecommand{\urlprefix}{URL }
\providecommand{\doi}[1]{https://doi.org/#1}

\bibitem{ahe+09}
Achanta, R., Hemami, S., Estrada, F., Susstrunk, S.: Frequency-tuned salient region detection. In: 2009 IEEE conference on computer vision and pattern recognition (2009)

\bibitem{agb+21}
Amir, S., Gandelsman, Y., Bagon, S., Dekel, T.: Deep vit features as dense visual descriptors. arXiv preprint arXiv:2112.05814  (2021)

\bibitem{b00}
Bradski, G.: {The OpenCV Library}. Dr. Dobb's Journal of Software Tools  (2000)

\bibitem{bc15}
Bui, T., Collomosse, J.: Scalable sketch-based image retrieval using color gradient features. In: ICCV (2015)

\bibitem{brp+16b}
Bui, T., Ribeiro, L., Ponti, M., Collomosse, J.: Generalisation and sharing in triplet convnets for sketch based visual search. In: arXiv (2016)

\bibitem{ctm+21}
Caron, M., Touvron, H., Misra, I., J{\'e}gou, H., Mairal, J., Bojanowski, P., Joulin, A.: Emerging properties in self-supervised vision transformers. In: Proceedings of the IEEE/CVF International Conference on Computer Vision (2021)

\bibitem{cmh+14}
Cheng, M.M., Mitra, N.J., Huang, X., Hu, S.M.: {SalientShape}: Group saliency in image collections. Vis. Comput.  \textbf{30}(4),  443–453 (apr 2014)

\bibitem{cwl+13}
Cheng, M.M., Warrell, J., Lin, W.Y., Zheng, S., Vineet, V., Crook, N.: Efficient salient region detection with soft image abstraction. In: Proceedings of the IEEE International Conference on Computer vision (2013)

\bibitem{drd+19}
Dey, S., Riba, P., Dutta, A., Llados, J., Song, Y.Z.: Doodle to search: Practical zero-shot sketch-based image retrieval. In: CVPR (2019)

\bibitem{dz13}
Doll{\'a}r, P., Zitnick, C.L.: Structured forests for fast edge detection. In: ICCV (2013)

\bibitem{dsl+09}
Dong, W., Socher, R., Li-Jia, L., Li, K., Fei-Fei, L.: {ImageNet}: A large-scale hierarchical image database. In: CVPR (2009)

\bibitem{dbk+20}
Dosovitskiy, A., Beyer, L., Kolesnikov, A., Weissenborn, D., Zhai, X., Unterthiner, T., Dehghani, M., Minderer, M., Heigold, G., Gelly, S., et~al.: An image is worth 16x16 words: Transformers for image recognition at scale. arXiv preprint arXiv:2010.11929  (2020)

\bibitem{etc22}
Efthymiadis, N., Tolias, G., Chum, O.: Edge augmentation for large-scale sketch recognition without sketches. arXiv preprint arXiv:2202.13164  (2022)

\bibitem{evw+10}
Everingham, M., Van~Gool, L., Williams, C.K.I., Winn, J., Zisserman, A.: The {PASCAL} visual object classes {(VOC)} challenge. IJCV  \textbf{88}(2),  303--338 (Jun 2010)

\bibitem{fi13}
Faktor, A., Irani, M.: Co-segmentation by composition. In: Proceedings of the IEEE international conference on computer vision (2013)

\bibitem{fcl+17}
Fan, D.P., Cheng, M.M., Liu, Y., Li, T., Borji, A.: Structure-measure: A new way to evaluate foreground maps. In: Proceedings of the IEEE international conference on computer vision (2017)

\bibitem{flj20}
Fan, D.P., Lin, Z., Ji, G.P., Zhang, D., Fu, H., Cheng, M.M.: Taking a deeper look at co-salient object detection. In: Proceedings of the IEEE/CVF conference on computer vision and pattern recognition (2020)

\bibitem{fff+21}
Fan, Q., Fan, D.P., Fu, H., Tang, C.K., Shao, L., Tai, Y.W.: Group collaborative learning for co-salient object detection. In: Proceedings of the IEEE/CVF Conference on Computer Vision and Pattern Recognition (2021)

\bibitem{jsm+19}
He, J., Zhang, S., Yang, M., Shan, Y., Huang, T.: Bi-directional cascade network for perceptual edge detection. In: CVPR (2019)

\bibitem{hlc+18}
Hsu, K.J., Lin, Y.Y., Chuang, Y.Y., et~al.: Co-attention cnns for unsupervised object co-segmentation. In: IJCAI (2018)

\bibitem{htl+18}
Hsu, K.J., Tsai, C.C., Lin, Y.Y., Qian, X., Chuang, Y.Y.: Unsupervised cnn-based co-saliency detection with graphical optimization. In: Proceedings of the European Conference on Computer Vision (ECCV) (2018)

\bibitem{rc13}
Hu, R., Collomosse, J.: A performance evaluation of gradient field hog descriptor for sketch based image retrieval. CVIU  (2013)

\bibitem{jgs10}
Jacobs, D.E., Goldman, D.B., Shechtman, E.: Cosaliency: Where people look when comparing images. In: Proc. UIST (2010)

\bibitem{kenton2019bert}
Kenton, J.D.M.W.C., Toutanova, L.K.: Bert: Pre-training of deep bidirectional transformers for language understanding. In: Proceedings of naacL-HLT (2019)

\bibitem{SelfReg}
Kim, D., Yoo, Y., Park, S., Kim, J., Lee, J.: Selfreg: Self-supervised contrastive regularization for domain generalization. In: ICCV (2021)

\bibitem{lsl+19}
Li, B., Sun, Z., Li, Q., Wu, Y., Hu, A.: Group-wise deep object co-segmentation with co-attention recurrent neural network. In: Proceedings of the IEEE/CVF International Conference on Computer Vision (2019)

\bibitem{lsw+19}
Li, B., Sun, Z., Wang, Q., Li, Q.: Co-saliency detection based on hierarchical consistency. In: Proceedings of the 27th ACM International Conference on Multimedia (2019)

\bibitem{lhr19}
Li, W., Hosseini~Jafari, O., Rother, C.: Deep object co-segmentation. In: Computer Vision--ACCV 2018: 14th Asian Conference on Computer Vision, Perth, Australia, December 2--6, 2018, Revised Selected Papers, Part III 14 (2019)

\bibitem{mvm11}
Maji, S., Vishnoi, N.K., Malik, J.: Biased normalized cuts. In: CVPR 2011. IEEE (2011)

\bibitem{mrl+22}
Melas-Kyriazi, L., Rupprecht, C., Laina, I., Vedaldi, A.: Deep spectral methods: A surprisingly strong baseline for unsupervised semantic segmentation and localization. In: Proceedings of the IEEE/CVF Conference on Computer Vision and Pattern Recognition (2022)

\bibitem{SagNet}
Nam, H., Lee, H., Park, J., Yoon, W., Yoo, D.: Reducing domain gap by reducing style bias. In: CVPR (2021)

\bibitem{pm14}
Parui, S., Mittal, A.: Similarity-invariant sketch-based image retrieval in large databases. In: ECCV (2014)

\bibitem{qhz+16}
Quan, R., Han, J., Zhang, D., Nie, F.: Object co-segmentation via graph optimized-flexible manifold ranking. In: Proceedings of the IEEE conference on computer vision and pattern recognition (2016)

\bibitem{rtc18}
Radenovic, F., Tolias, G., Chum, O.: Deep shape matching. In: ECCV (2018)

\bibitem{rkb04}
Rother, C., Kolmogorov, V., Blake, A.: " grabcut" interactive foreground extraction using iterated graph cuts. ACM transactions on graphics (TOG)  (2004)

\bibitem{rother2006cosegmentation}
Rother, C., Minka, T., Blake, A., Kolmogorov, V.: Cosegmentation of image pairs by histogram matching-incorporating a global constraint into mrfs. In: 2006 IEEE Computer Society Conference on Computer Vision and Pattern Recognition (CVPR'06) (2006)

\bibitem{sbo15}
Saavedra, J.M., Barrios, J.M., Orand, S.: Sketch based image retrieval using learned keyshapes ({LKS}). In: BMVC (2015)

\bibitem{sbh+16}
Sangkloy, P., Burnell, N., Ham, C., Hays, J.: The sketchy database: learning to retrieve badly drawn bunnies. ACM Transactions on Graphics  (2016)

\bibitem{sdm17}
Seddati, O., Dupont, S., Mahmoudi, S.: Quadruplet networks for sketch-based image retrieval. In: ICMR (2017)

\bibitem{shi+97}
Shi, J., Malik, J.: Normalized cuts and image segmentation. In: CVPR (1997)

\bibitem{sma97}
Shi, J., Malik, J.: Normalized cuts and image segmentation. In: CVPR (Jun 1997)

\bibitem{sax22}
Shin, G., Albanie, S., Xie, W.: Unsupervised salient object detection with spectral cluster voting. In: Proceedings of the IEEE/CVF Conference on Computer Vision and Pattern Recognition. pp. 3971--3980 (2022)

\bibitem{sxa22}
Shin, G., Xie, W., Albanie, S.: Reco: Retrieve and co-segment for zero-shot transfer. Advances in Neural Information Processing Systems  \textbf{35},  33754--33767 (2022)

\bibitem{spv+21}
Sim{\'e}oni, O., Puy, G., Vo, H.V., Roburin, S., Gidaris, S., Bursuc, A., P{\'e}rez, P., Marlet, R., Ponce, J.: Localizing objects with self-supervised transformers and no labels. In: BMVC-British Machine Vision Conference (2021)

\bibitem{sds+22}
Su, Y., Deng, J., Sun, R., Lin, G., Wu, Q.: A unified transformer framework for group-based segmentation: Co-segmentation, co-saliency detection and video salient object detection. arXiv preprint arXiv:2203.04708  (2022)

\bibitem{tc17}
Tolias, G., Chum, O.: Asymmetric feature maps with application to sketch based retrieval. In: CVPR (2017)

\bibitem{wly+19}
Wang, X., Liang, X., Yang, B., Li, F.W.B.: No-reference synthetic image quality assessment with convolutional neural network and local image saliency. Computational visual media.  \textbf{5}(2) (June 2019)

\bibitem{wsh+22}
Wang, Y., Shen, X., Hu, S.X., Yuan, Y., Crowley, J.L., Vaufreydaz, D.: Self-supervised transformers for unsupervised object discovery using normalized cut. In: Proceedings of the IEEE/CVF Conference on Computer Vision and Pattern Recognition (2022)

\bibitem{Wlq+21}
Wang, Z., Luo, Y., Qiu, R., Huang, Z., Baktashmotlagh, M.: Learning to diversify for single domain generalization. In: ICCV (2021)

\bibitem{wzw+19}
Wei, X.S., Zhang, C.L., Wu, J., Shen, C., Zhou, Z.H.: Unsupervised object discovery and co-localization by deep descriptor transformation. Pattern Recognition  (2019)

\bibitem{sz17}
Xie, S., Tu, Z.: Holistically-nested edge detection. IJCV  (2017)

\bibitem{yls+16}
Yu, Q., Lie, F., Song, Y.Z., Xian, T., Hospedales, T., Loy, C.C.: Sketch me that shoe. In: CVPR (2016)

\bibitem{yxz+22}
Yu, S., Xiao, J., Zhang, B., Lim, E.G.: Democracy does matter: Comprehensive feature mining for co-salient object detection. In: Proceedings of the IEEE/CVF Conference on Computer Vision and Pattern Recognition (2022)

\bibitem{zll+21}
Zhang, C., Li, G., Lin, G., Wu, Q., Yao, R.: Cyclesegnet: Object co-segmentation with cycle refinement and region correspondence. IEEE Transactions on Image Processing  (2021)

\bibitem{zhl+16}
Zhang, D., Han, J., Li, C., Wang, J., Li, X.: Detection of co-salient objects by looking deep and wide. International Journal of Computer Vision  (2016)

\bibitem{zcl+20}
Zhang, K., Chen, J., Liu, B., Liu, Q.: Deep object co-segmentation via spatial-semantic network modulation. In: Proceedings of the AAAI Conference on Artificial Intelligence (2020)

\bibitem{zzl+19}
Zhang, L., Zhang, J., Lin, Z., Lu, H., He, Y.: {CapSal}: Leveraging captioning to boost semantics for salient object detection. In: CVPR (June 2019)

\bibitem{zhl+21}
Zhang, N., Han, J., Liu, N., Shao, L.: Summarize and search: Learning consensus-aware dynamic convolution for co-saliency detection. In: Proceedings of the IEEE/CVF International Conference on Computer Vision (2021)

\bibitem{zjx+20}
Zhang, Z., Jin, W., Xu, J., Cheng, M.M.: Gradient-induced co-saliency detection. In: European Conference on Computer Vision (ECCV) (2020)

\end{thebibliography}

\newpage


\appendix
\section{Appendix}

The Appendix includes qualitative and quantitative results of the proposed co-segmentation method on samples from classes that are not part of ImageNet and the implementation details of the proposed method.

\subsection{Out-of-ImageNet classes}
\begin{figure}[ht]
\centering
\begin{tabular}{ccc}
Class Name & Sample RGB & Predicted mask \\
\hline \\
\raisebox{0.75cm}{chopsticks}&
\includegraphics[height=1.6cm, width=1.6cm]{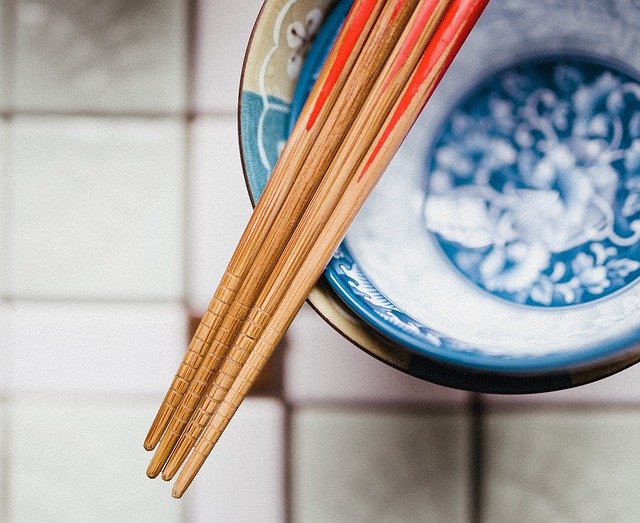}&
\includegraphics[height=1.6cm, width=1.6cm]{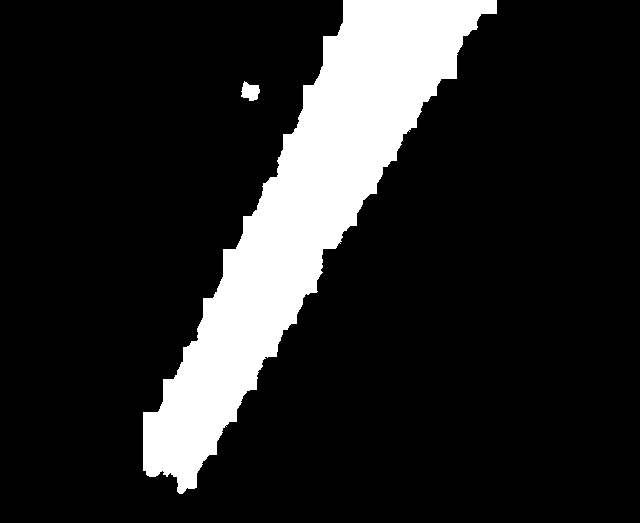}
\\
\raisebox{0.75cm}{macaroon}&
\includegraphics[height=1.6cm, width=1.6cm]{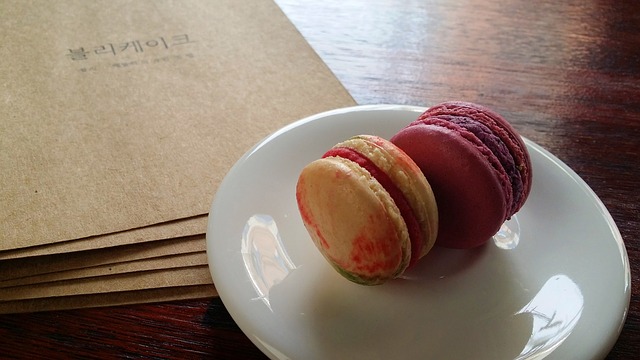}&
\includegraphics[height=1.6cm, width=1.6cm]{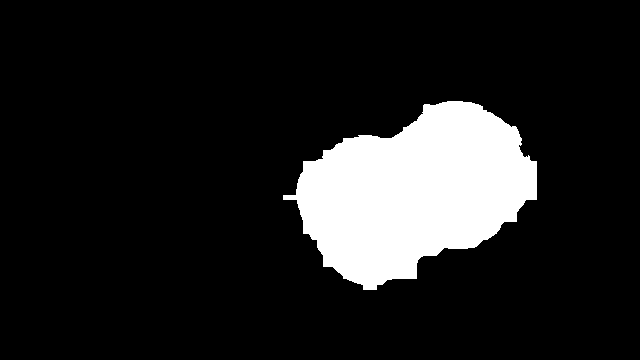}
\\
\raisebox{0.75cm}{clover}&
\includegraphics[height=1.6cm, width=1.6cm]{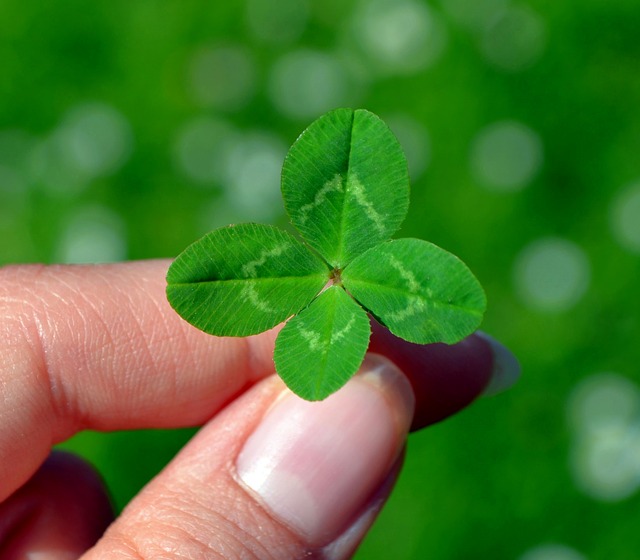}&
\includegraphics[height=1.6cm, width=1.6cm]{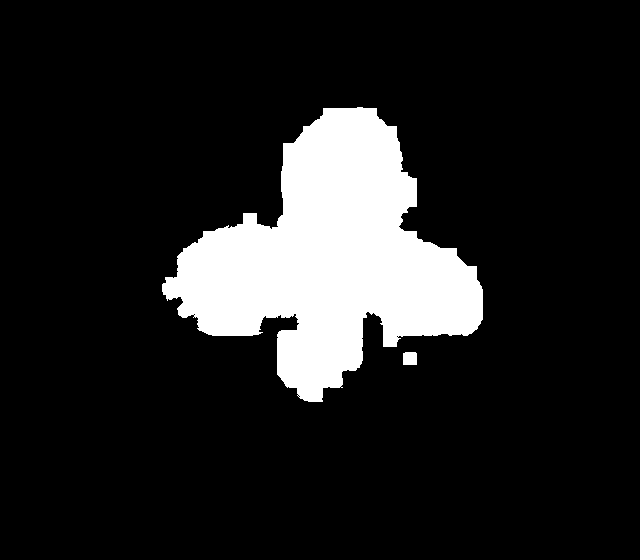}
\\
\raisebox{0.75cm}{persimmon} &
\includegraphics[height=1.6cm, width=1.6cm]{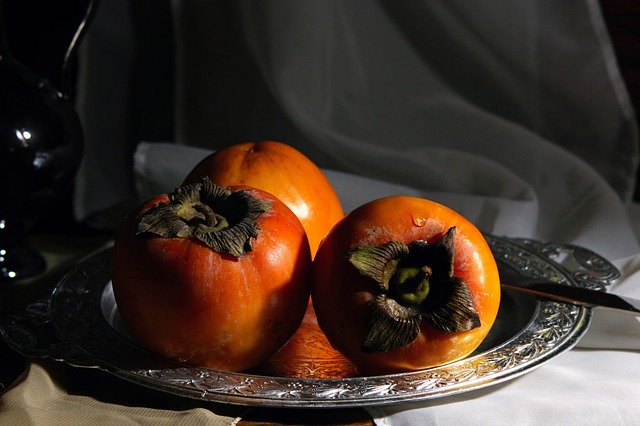}&
\includegraphics[height=1.6cm, width=1.6cm]{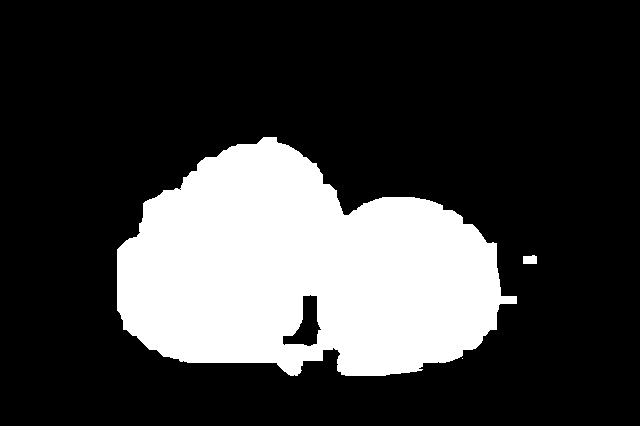}
\\
\raisebox{0.75cm}{rocking horse} &
\includegraphics[height=1.6cm, width=1.6cm]{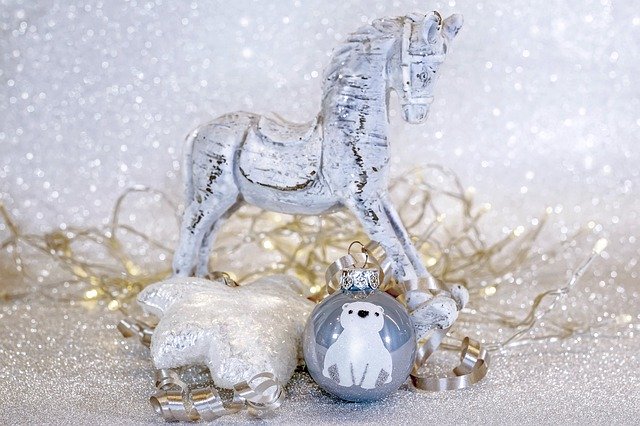}&
\includegraphics[height=1.6cm, width=1.6cm]{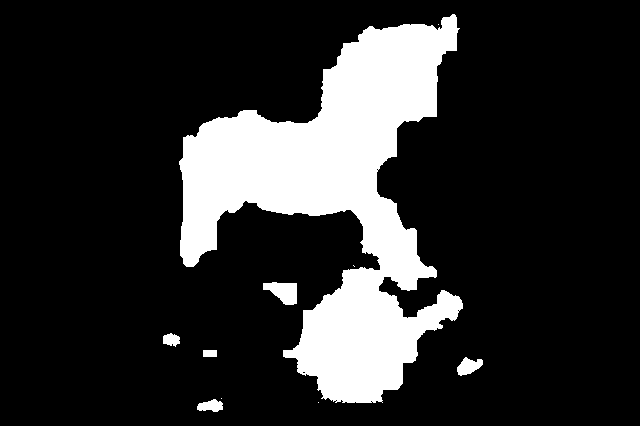}
\\
\end{tabular}
\caption{
Examples of co-segmentation for classes that are not part of the ImageNet dataset, which coincides with the pretraining dataset of the backbones utilized by our method.
}
\label{fig:non_class}
\end{figure}

\begin{figure}[ht]
\centering
\begin{tabular}{ccc}
Class Name & Sample RGB & Predicted mask \\
\hline \\
\raisebox{0.75cm}{chopsticks}&
\includegraphics[height=1.6cm, width=1.6cm]{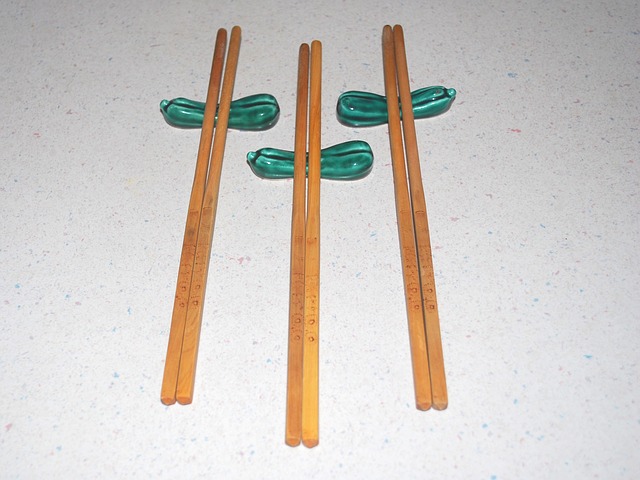}&
\includegraphics[height=1.6cm, width=1.6cm]{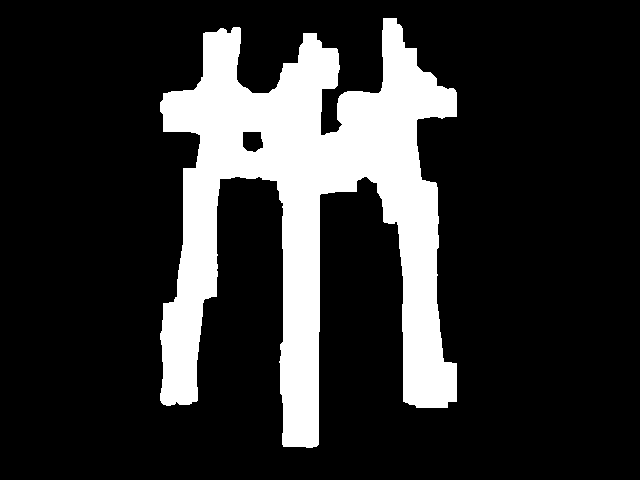}
\\
\raisebox{0.75cm}{pinecone}&
\includegraphics[height=1.6cm, width=1.6cm]{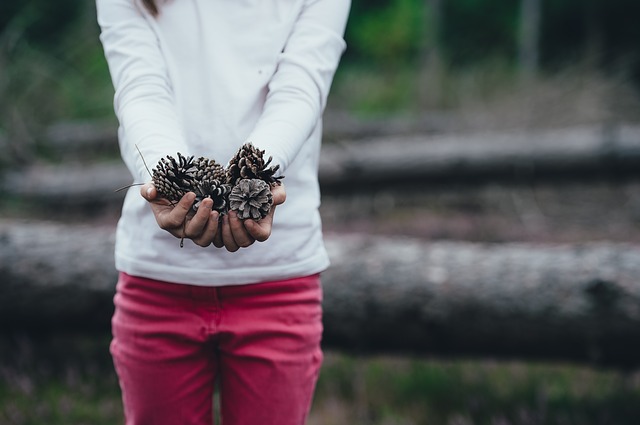}&
\includegraphics[height=1.6cm, width=1.6cm]{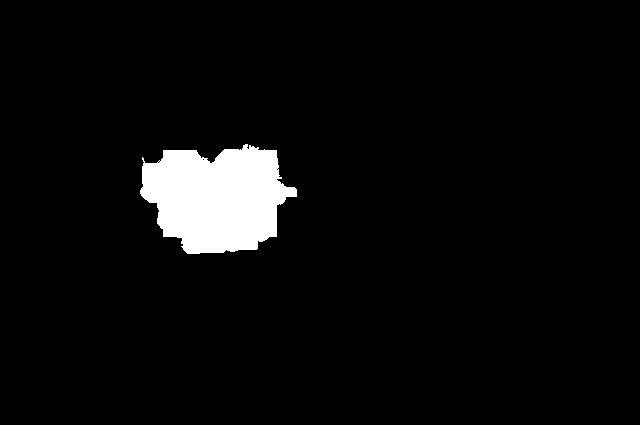}
\\
\raisebox{0.75cm}{clover}&
\includegraphics[height=1.6cm, width=1.6cm]{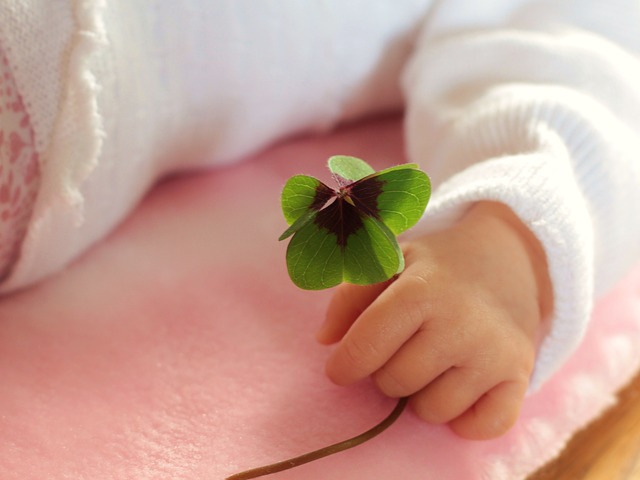}&
\includegraphics[height=1.6cm, width=1.6cm]{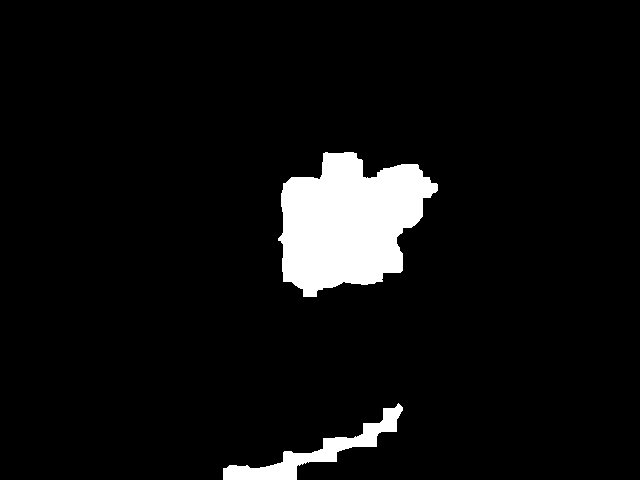}
\\
\raisebox{0.75cm}{persimmon} &
\includegraphics[height=1.6cm, width=1.6cm]{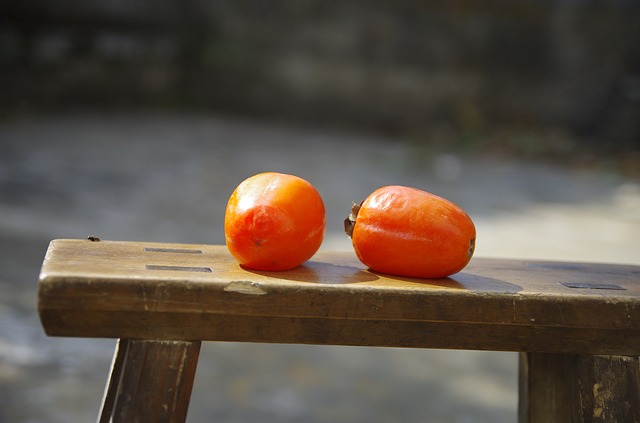}&
\includegraphics[height=1.6cm, width=1.6cm]{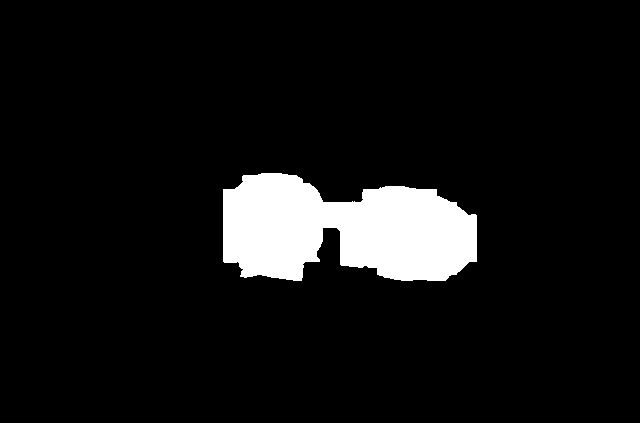}
\\
\raisebox{0.75cm}{rocking horse}&
\includegraphics[height=1.6cm, width=1.6cm]{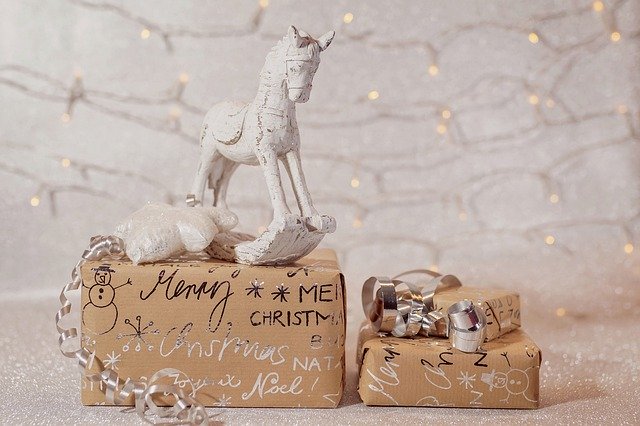}&
\includegraphics[height=1.6cm, width=1.6cm]{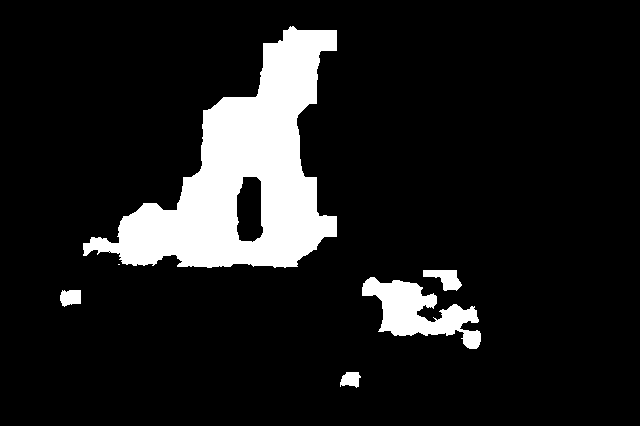}
\\
\raisebox{0.75cm}{high heels} &
\includegraphics[height=1.6cm, width=1.6cm]{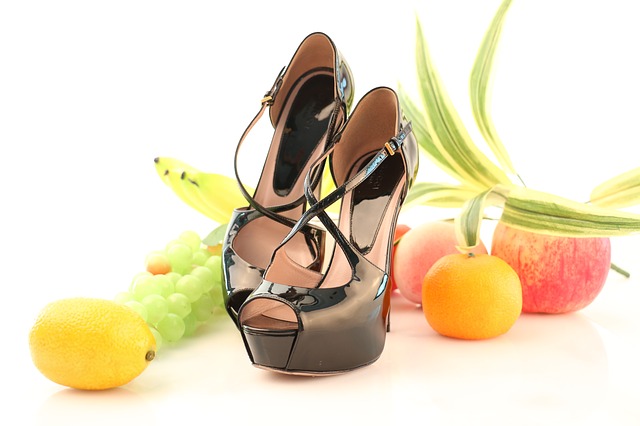}&
\includegraphics[height=1.6cm, width=1.6cm]{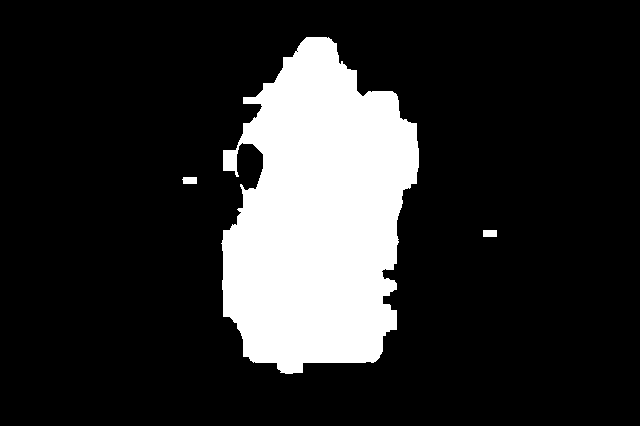}
\\
\raisebox{0.75cm}{dice}&
\includegraphics[height=1.6cm, width=1.6cm]{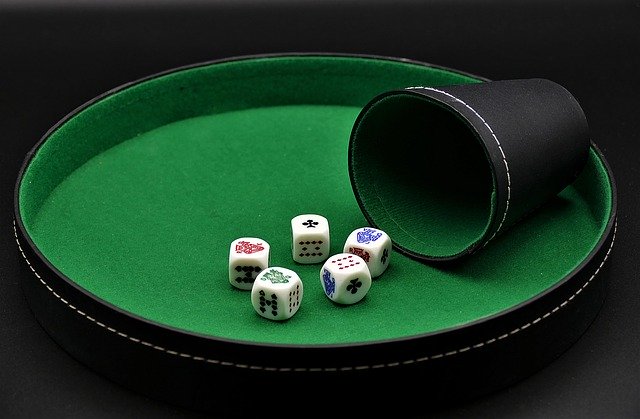}&
\includegraphics[height=1.6cm, width=1.6cm]{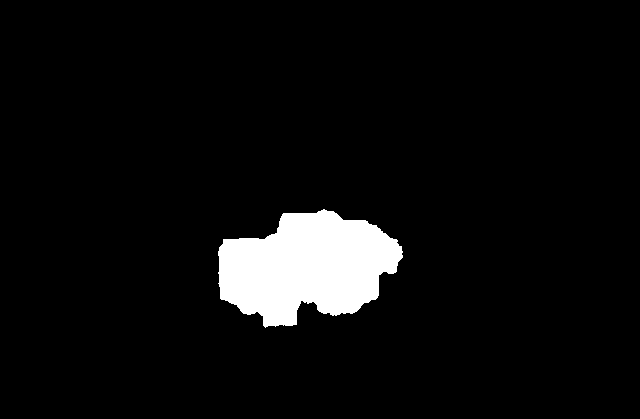}
\\

\end{tabular}
\caption{
Examples of co-segmentation for classes that are not part of the ImageNet dataset, which coincide with the pretraining dataset of the backbones utilized by our method.
}
\label{fig:non_class2}
\end{figure}

We include additional examples of co-salient object detection on the CoCA dataset from classes that are not part of the ImageNet dataset.
That is to show qualitatively how the proposed method works on classes outside of the pretraining data of the backbones it utilizes.
The examples are shown in~\cref{fig:non_class,fig:non_class2}.

We perform additional experiments to support the claim quantitatively. 
First, we split the CoCA dataset into ImageNet and non-ImageNet classes (21 and 59, respectively). This was achieved by comparing the class labels with BERT~\cite{kenton2019bert} embeddings and manually verifying the matches. The results evaluated on the non-ImageNet part of the dataset, comparing the proposed method and the baselines, are shown in Table~\ref{tab:suppl_unseen_IN}.

A generalization to unseen objects can also be interpreted as how well the co-segmentation works on images that are not used to build the model. 
To this end, we performed an additional leave-one-out experiment. 
Each image in the CoCA dataset is segmented by a class-relevance model that is obtained from all other class examples (excluding the tested image). 
The overall performance is not affected: 0.124 $MAE$, 0.531 $F_{\beta}^{max}$, 0.670 $S_{\alpha}$. This experiment demonstrates generalization beyond training images.

\begin{table}[ht]
  \centering
  \caption{
  }
    \scalebox{0.85}{
    \begin{tabular}{c|ccc}
    \hline
    &  \multicolumn{3}{|c}{non-ImageNet CoCA}\\ 
    \hline
    Method & $MAE \downarrow$ & $ F_{\beta}^{\max}\uparrow$ &$S_{\alpha}\uparrow$ \\
    \hline
    Amir et al. [2]  & 0.253 & 0.391 & 0.550\\
    N-cut [46]  & 0.145 & 0.478 & 0.635\\
    CBNC (ours)  & {\bf 0.128} & {\bf 0.525} & {\bf 0.666}\\
    \hline
    \end{tabular}
    }

  \label{tab:suppl_unseen_IN}
\end{table}

\subsection{Implementation details and timings}
For all co-segmentations, images are resized to 256$\times$256 pixels.
We use DINO ViT-Small/8~\cite{ctm+21} and ImageNet ViT-Small/16~\cite{dbk+20} (the latter is not publicly available for the patch size of 8).
For the Im4Sketch dataset of the sketch classification application, we use 90 images per class to calculate $\xi$ in equation (1) of the main paper, as we have observed that using more shows no advantage.
We set the hyperparameter $\tau = 0.2$ similar to~\cite{wsh+22}, $\gamma = 10^{-4}$, $\epsilon = 10^{-5}$ and use 16 eigenvectors to form the biased N-cut vector.
The value of the temperature $\beta$ in the softmax is set to 0.5.

The time needed to extract the ViT features for a set of 90 images is approximately 4 seconds on an NVIDIA GTX TITAN X GPU.
Calculating the first eigenvector on ViT features from 90 images for class relevance takes around 0.7 seconds on an Intel Xeon E5-2620 v3 CPU.
The estimation of the mask by the biased N-cut takes approximately 0.8 seconds for each image on the same CPU.

\end{document}